\pdfoutput=1
\documentclass[11pt]{article}

\usepackage{amsmath, amssymb, amsfonts}
\usepackage{graphicx}
\usepackage{hyperref}
\usepackage{authblk}   
\usepackage{geometry}
\usepackage{cleveref}
\usepackage{algorithm}
\usepackage{algorithmic}

\hypersetup{
	colorlinks=true,
	linkcolor=blue,
	citecolor=blue,
	urlcolor=blue
}

\title{LHMCF-Net: A Learned Hyperbolic Mean Curvature Flow Network for Medical Images Segmentation}

\author[1]{Shuangshuang Duan}
\author[2]{Chunlei He}
\author[2]{Shoujun Huang}
\author[3]{Dexing Kong}

\affil[1]{Zhejiang Normal University, Jinhua 321004 China}
\affil[2]{College of Mathematical Medicine, Zhejiang Normal University}
\affil[2]{School of Mathematical Sciences, Zhejiang University, Hangzhou 310027 China}

\date{}  

\begin{document}
	
	\maketitle
	
	\begin{abstract}
		Motivated by the classical Chan-Vese model and the ability of deep priors to capture complex spatial structures, we develop a segmentation model that leverages learned hyperbolic mean curvature flow (LHMCF) as a mathematical foundation for integrating feature space data fidelity and deep structural priors within a unified high‑dimensional framework. The proposed LHMCF model is governed by a second‑order dissipative hyperbolic PDE, where the introduction of a velocity field provides inertia and momentum to the evolving interface. This hyperbolic mechanism enables the contour to bypass noise‑induced local minima and propagate coherently through low‑contrast or ambiguous regions, addressing limitations inherent to first‑order parabolic flows. To solve the continuous LHMCF model, we construct a deep unfolding network, named LHMCF‑Net, which maps the iterative numerical procedure of the PDE into a sequence of discrete evolution stages. Each stage corresponds to one physically interpretable update of the underlying dynamical system, allowing the network to inherit the stability and geometric consistency of the PDE while supporting end‑to‑end optimization. Comprehensive experiments on three publicly available medical segmentation datasets demonstrate that LHMCF‑Net achieves superior performance, particularly in challenging scenarios with low contrast and unclear boundaries. These results highlight the effectiveness of embedding hyperbolic geometric evolution into deep unfolding architectures and underscore the potential of physically inspired models for robust medical image segmentation.
	\end{abstract}
	
	\noindent\textbf{Keywords:} image segmentation, hyperbolic mean curvature flow, deep unfolding network, level-set evolution
	
	\section{Introduction}
	Image segmentation is a fundamental task in computer vision and plays a crucial role in medical diagnostics, clinical research, and pathological analysis \cite{chen2021deep, LITJENS201760survey, pham2000current}. Medical images, however, are often degraded by noise and artifacts due to equipment limitations and tissue characteristics. For example, ultrasound images typically contain strong speckle noise and shadow artifacts \cite{Noble2026ultrasound}. In addition, medical images usually exhibit low contrast, with lesion regions sharing similar color and texture with surrounding tissues \cite{Dong2021Polyp-PVT}. These factors make it difficult to delineate lesion boundaries and shapes, rendering accurate segmentation a particularly challenging problem.
	
	Traditional image segmentation methods are often formulated within variational frameworks, where the task is posed as minimizing an energy functional composed of a fidelity term and a regularization term \cite{huang2008fast, li2007implicit, Li2010Distance, lorigo2001curves, morel2012variational, vese2002multiphase, yang2019image}. The fidelity term enforces consistency with the observed data, while the regularization term encodes prior assumptions about the geometry or smoothness of the segmented regions. Representative examples include the snake model \cite{kass1988snakes}, Chan-Vese \cite{Chan2001active}, Mumford-Shah \cite{mumford1989optimal},  geodesic active contours \cite{caselles1997geodesic}, and so on. Although these model-based methods have strong mathematical foundations, they operate primarily on image scale space representations and do not incorporate high-dimensional features learned from data. Consequently, their performance may be limited for medical images with low contrast, weak boundaries, or missing edges. In addition, segmentation accuracy is sensitive to the choice of regularization terms, which often must be manually designed for different imaging modalities and tasks. This reduces model flexibility and makes it difficult to construct regularization terms suitable for complex segmentation scenarios. Extensive parameter tuning further increases the computational and practical burden.
	
	In recent years, the rapid growth of computational resources has enabled data‑driven deep neural networks to become highly effective tools for image segmentation. Early convolutional architectures, such as U-Net \cite{Ronneberger2015Unet}, demonstrated that encoder–decoder designs with skip connections can recover fine‑scale spatial information while capturing high-level semantics. Subsequent developments focused on enhancing multi‑scale context modeling and boundary accuracy. Representative examples include pyramid‑based modules \cite{zhao2017pyramid} and atrous convolution strategies \cite{chen2018deeplab}, as used in the DeepLab family, which significantly improved performance on objects with varying spatial scales. More recently, vision transformers (ViTs) \cite{arnab2021vivit} have attracted considerable attention due to their ability to model long‑range dependencies and capture global contextual information. Transformer-based components have therefore been integrated into segmentation networks to complement or replace convolutional modules. For example, TransUNet \cite{chen2021transunet} combines a ViT encoder with a U‑Net style decoder, leveraging transformer representations for global feature modeling while retaining convolutional operations for spatial reconstruction. These networks have achieved state‑of‑the‑art performance across a wide range of segmentation tasks and often outperform traditional model‑based methods when sufficient annotated data are available. Nevertheless, the outputs of deep neural networks are difficult to analyze from a mathematical perspective, and the structural prior of the semantic regions is not fully utilized.
	
	A natural way to address these limitations is to integrate the strengths of variational methods and deep neural networks for image segmentation tasks. Variational models provide a clear mathematical framework, well‑defined energy formulations, and interpretable geometric priors, while deep learning offers strong representational capacity and robustness to noise. Recent studies have explored connections between model-based models and neural networks. One direct strategy is to incorporate functions derived from partial differential equation (PDE) models into the loss function of deep networks \cite{chen2019learning, kim2019mumford, kim2019cnn}. In this setting, the classical energy functional or PDE residual is added as a regularization term, so that the training objective enforces both data fidelity and structural priors. This approach enhances interpretability and robustness, but requires careful balancing of loss weights and often increases computational cost. Another prominent direction is the deep unfolding framework, which constructs network architectures based on the iterative schemes used to solve mathematical models \cite{le2024unfolded, lunz2018adversarial, ren2025DecNet, zhang2025lms}. In classical variational segmentation, minimizing the energy functional leads to either a gradient descent PDE or an iterative optimization algorithm, traditionally solved by finite difference or finite element methods. Deep unfolding instead takes the iterative update as the backbone of a neural architecture, expands the iterations into network layers, and trains the parameters in a data-driven manner. This approach preserves the mathematical structure of the original model while leveraging the flexibility of deep learning.
	
	In this paper, we introduce a novel segmentation model that integrates feature-space data fidelity and deep structural priors into a hyperbolic mean curvature flow framework. By introducing a velocity field to complement the level-set function, the second-order formulation simulates a damped physical system characterized by inertia and momentum. Compared with traditional first-order parabolic flows that evolve purely through gradient descent, the hyperbolic dynamics enable the interface to traverse noise‑induced local minima and propagate coherently through ambiguous regions. To realize this continuous model computationally, we employ a deep unfolding strategy that discretizes the coupled PDE system into a sequence of explicit evolution stages. At each stage, we alternately update the velocity field, the level‑set function, and the foreground/background feature means. In particular, the foreground feature means are updated using a momentum‑based rule, while the background means follow an exponential moving average. By embedding the constraint $|\nabla\phi|=1$ directly into the hybrid loss, we ensure that the level‑set function maintains the signed‑distance property throughout the unfolding process, eliminating the need for manual re‑initialization and preserving the topological integrity of the evolving interface. 
	
	The main contributions of this work are the following.
	\begin{itemize}
		\item We propose the LHMCF model, which integrates curvature-driven smoothing, feature space data fidelity, and the deep structural prior within a unified framework. This formulation bridges classical geometric evolution with modern semantic representation.
		\item Our model incorporates both an acceleration term and a dissipative damping term. The acceleration provides inertia that helps the interface escape shallow local minima, while the damping stabilizes the evolution and prevents oscillatory behavior. Their combined effect yields a fast, robust, and noise‑resistant segmentation process that fundamentally differs from first‑order parabolic flows.
		\item We reformulate the coupled hyperbolic system into a sequence of learnable unfolding stages, where each stage corresponds to a physically meaningful update of the underlying dynamics. This architecture not only ensures numerical stability, but also provides a transparent interpretation of how the learned parameters influence the evolution, offering a  connection between continuous PDE dynamics and discrete neural operators.
	\end{itemize}
	
	The remainder of this paper is organized as follows. In \cref{sec:related}, we review some most recently related works. The proposed LHMCF segmentation model is presented in \cref{sec:method}. The experimental results are reported in \cref{sec:experiments}. In \cref{sec:ablation}, we provide the ablation analysis and physical interpretability analyses of our model. Finally, our conclusions and discussions are given in \cref{sec:conclusion}.
	
	\section{Related works}
	\label{sec:related}
	\subsection{CV Model}
	The Chan-Vese (CV) model \cite{Chan2001active} is a representative region‑based variational method for image segmentation derived from a piecewise constant form of the Mumford–Shah functional. It seeks a contour $C$ that partitions the image domain $\Omega$ into two regions with approximately homogeneous intensities. For a given input image $I(x,y)$, the energy functional is defined as
	\begin{displaymath}
		E(c_1, c_2, C)=\mu \cdot Length(C) + \lambda_1 \int _{inside(C)}|I(x,y)-c_1|^2dxdy + \lambda_2 \int _{outside(C)}|I(x,y)-c_2|^2dxdy,
	\end{displaymath}
	where $\mu \geq 0$, $\lambda_1, \lambda_2>0$ are fixed parameters, and $c_1$, $c_2$ denote the mean intensities inside and outside $C$, respectively. The first term is the regularization term that controls the length of the evolving contour. The second and third terms are the fidelity terms, which ensure that $C$ does not deviate significantly from the actual object boundary. Unlike edge‑based active contours, the CV model can segment objects with weak or missing gradients because it relies on region statistics rather than local image gradients.
	
	To handle topology changes, such as region splitting and merging, the level set method \cite{Osher1988fronts} is generally adopted. Under this framework, the contour $C$ is represented by the zero level set of a Lipschitz function $\phi:\Omega\rightarrow \mathbb{R}$,
	\begin{displaymath}
		\left\{\begin{array}{ll}
			C = \left\{(x,y)\in \Omega: \phi(x,y)=0 \right\},  \vspace{2mm}\\
			inside(C) = \left\{(x,y)\in \Omega: \phi(x,y)>0\right\},   \vspace{2mm}\\
			outside(C) = \left\{(x,y)\in \Omega: \phi(x,y)<0\right\}.
		\end{array}\right.
	\end{displaymath}
	With the regularized Heaviside function $H_{\epsilon}(\phi)$, and the Dirac delta approximation$\delta_{\epsilon}(\phi)$, the energy functional can be written in the level set form
	\begin{align*}
		E(c_1, c_2, \phi) &= \mu \int_{\Omega} \delta_{\epsilon}(\phi(x,y))|\nabla \phi(x,y)|dxdy\\
		&+ \lambda_1 \int_{\Omega} |I(x,y)-c_1|^2 H_{\epsilon}(\phi(x,y))dxdy
		+ \lambda_2 \int_{\Omega}|I(x,y)-c_2|^2(1-H_{\epsilon}(\phi(x,y))dxdy,
	\end{align*}
	where
	\begin{displaymath}
		H_{\epsilon}(\phi)
		= \frac{1}{2}\Big(1 + \frac{2}{\pi}\arctan\Big(\frac{\phi}{\varepsilon}\Big)
		\Big), \;
		\delta_{\epsilon}(\phi)
		= \frac{1}{\pi}\cdot\frac{\varepsilon}{\phi^2 + \varepsilon^2}.
	\end{displaymath}
	At each iteration, the optimal values of $c_1$ and $c_2$  are updated explicitly by
	\begin{displaymath}
		c_1 = \frac{\int_{\Omega} I(x,y)H_{\epsilon}(\phi(x,y))dxdy}{\int_{\Omega}H_{\epsilon}(\phi(x,y))dxdy},\quad
		c_2 = \frac{\int_{\Omega} I(x,y)(1-H_{\epsilon}(\phi(x,y)))dxdy}{\int_{\Omega}(1-H_{\epsilon}(\phi(x,y)))dxdy}.
	\end{displaymath}
	Fixing $c_1$ and $c_2$, minimizing the energy leads to the Euler-Lagrange equation
	\begin{displaymath}
		\frac{\partial \phi}{\partial t}=\delta_{\varepsilon}(\phi)\left[\mu\cdot div\left(\frac{\nabla\phi}{|\nabla\phi|}\right)
		-\lambda_1(I(x,y)-c_1)^2+\lambda_2(I(x,y)-c_2)^2\right].
	\end{displaymath}
	In the level set framework, the evolution equation is a parabolic PDE whose geometric term acts as a mean curvature flow, providing contour regularization. The region‑based force further drives the interface toward the boundary that best separates the two intensity regions, resulting in a mean curvature flow-like motion that stabilizes at the desired object boundary. In practice, the numerical implementation alternates between updating the region statistics and evolving the level set function, forming a stable and efficient optimization procedure. Due to its simplicity and robustness, the CV model has been widely used in medical image segmentation, particularly in settings where intensity distributions are relatively homogeneous within target regions.
	
	\subsection{Hyperbolic mean curvature flow}
	The hyperbolic mean curvature flow was originally introduced by Gurtin and Podio-Guidugli \cite{gurtin1991hyperbolic} to model melting-freezing waves on a crystal surface. In contrast to the classical mean curvature flow, which is a parabolic theory describing the velocity-driven evolution of plane curves and surfaces, Yau \cite{yau2000review} proposed replacing the velocity term $\frac{dX}{dt}$ with the acceleration $\frac{d^2X}{dt^2}$. This leads to the geometric evolution equation
	\begin{equation}\label{eq:21}
		\frac{d^2X}{dt^2}=\kappa\overrightarrow{N},
	\end{equation}
	where $\kappa$ is the mean curvature and $\overrightarrow{N}$ is the unit inner normal vector. He et al. \cite{he2009hyperbolic} established short-time existence and uniqueness of smooth solutions to \Cref{eq:21} and proved nonlinear stability in Euclidean spaces of dimension greater than four. For closed plane curves $\gamma:S^1\times[0,T)\rightarrow \mathbb{R}^2$, the one‑dimensional hyperbolic mean curvature flow \cite{dexing2009hyperbolic} takes the form
	\begin{displaymath}
		\frac{\partial ^2 \gamma}{\partial t^2}=\kappa\overrightarrow{N}
		-\left(\frac{\partial^2 \gamma}{\partial s \partial t}, \frac{\partial \gamma}{\partial t}\right)\overrightarrow{T},
	\end{displaymath}
	where $\overrightarrow{T}$ is the unit tangent vector and $s$ is the arclength parameter. Kong et al. \cite{dexing2009hyperbolic} proved that if the initial curve is a strictly convex closed curve, then there exists a class of initial velocities for which the solution exists only on a finite time interval $[0,T_{max})$. As $t\rightarrow T_{max}$, the evolving curve either converges to a point or develops shocks and other propagating discontinuities. An alternative dissipative hyperbolic mean curvature flow, as considered in \cite{wang2023dissipative}, is described by
	\begin{displaymath}
		\frac{\partial ^2 \gamma}{\partial t^2}+\beta\frac{\partial\gamma}{\partial t}=\kappa\overrightarrow{N}
		-\left(\frac{\partial^2 \gamma}{\partial s \partial t}, \frac{\partial \gamma}{\partial t}\right)\overrightarrow{T}
	\end{displaymath}
	where the damping parameter $\beta>0$. It can be shown that a nonnegative minimum initial velocity leads to convergence to a point or a limit curve, while a positive maximum initial velocity causes the flow to expand first and then converge to either a point or a limit curve. As regards the numerical approximation of hyperbolic mean curvature flow, Rotstein et al. \cite{rotstein1999hyperbolic} developed a hyperbolic crystalline algorithm for the motion of closed convex polygonal curves, which generalized the standard crystalline algorithm. Kusumasari \cite{kusumasari2018hyperbolic} considered the interface motion with an obstacle under hyperbolic mean curvature flow using a Hyperbolic Merriman-Bence-Osher (HMBO) algorithm. Deckelnick et al. \cite{Deckelnick2023Discrete} proposed a semidiscrete finite difference method for the approximation of hyperbolic mean curvature flow in the plane and derived error estimates in natural discrete norms. Monika \cite{suchomelova2025computational} analyzed the behavior of numerical solutions using a semi-discrete finite volume scheme. Duan et al. \cite{duan2026hyperbolic} employed physics‑informed neural networks to study the evolution of plane curves and surfaces governed by the hyperbolic mean curvature flow.
	
	Let us finally mention that second-order hyperbolic PDEs have recently been used for applications in image processing. Dong et al. \cite{dong2021class} studied the well-posedness of a class of second-order geometric quasilinear hyperbolic PDEs, and discussed their applications in denoising and displacement errors correction. Pau$\check{s}$ et al. \cite{pauvs2021segmentation} introduced an algorithm for segmentation of color images by means of the parametric mean curvature flow equation. Baravdish et al. \cite{baravdish2019damped} proposed a second-order nonlinear evolution equation for image denoising, demonstrating the potential of hyperbolic PDEs in image analysis.
	
	\subsection{Deep unfolding methods}
	The deep unfolding framework provides an efficient way to combine model‑based optimization with data‑driven learning. Its core idea is to translate iterative algorithms derived from variational or PDE‑based models into neural network architectures, where each iteration corresponds to a network layer. Linear operators such as filters or derivatives can be implemented by convolutional modules, and nonlinear steps can be approximated by suitable activation functions. In this way, deep unfolding retains the mathematical structure of classical models while benefiting from the flexibility of learned parameters. Several representative methods have demonstrated the effectiveness of this framework \cite{gregor2010learning, monga2021algorithm, xiang2021fista, zhang2018ista}. Yang et al. \cite{yang2018admm} introduced ADMM‑CSNet, which unrolls a generalized compressive sensing model and its Alternating Direction Method of Multipliers (ADMM) solver into a trainable architecture, enabling end‑to‑end learning of both model parameters and algorithmic updates for efficient reconstruction from sparse measurements. Jonas et al. \cite{adler2018learned} proposed the Learned Primal–Dual algorithm for tomographic reconstruction, where a primal–dual optimization scheme is unrolled, and the proximal operators are replaced by convolutional neural networks. Tai et al. \cite{tai2024pottsmgnet} developed PottsMGNet, which provides a mathematical interpretation of encoder–decoder architectures by reformulating segmentation as an optimal control problem based on the Potts model and discretizing it using a multigrid strategy with a hybrid operator‑splitting scheme.
	
	\section{The proposed method}
	\label{sec:method}
	\subsection{Learned hyperbolic mean curvature flow model}
	Suppose $\gamma: S^1\times[0,T)\rightarrow\mathbb{R}^2$ is a family of closed plane curves, where $S^1=[a,b]$ and $T>0$. We consider the following dissipative hyperbolic mean curvature flow
	\begin{equation}\label{eq:31}
		\frac{\partial ^2 \gamma}{\partial t^2}+\beta\frac{\partial\gamma}{\partial t}=\mu\kappa\overrightarrow{N}
		-\left(\frac{\partial^2 \gamma}{\partial s \partial t}, \frac{\partial \gamma}{\partial t}\right)\overrightarrow{T}, \quad \forall (u,t)\in S^1\times[0,T),
	\end{equation}
	with the initial conditions
	\begin{equation}\label{eq:32}
		\gamma(u,0) = \gamma_0, \; \gamma_t(u,0) = \gamma_1\overrightarrow{N}_0.
	\end{equation}
	Here $\beta>0$ is the damping parameter, $\kappa$ is the curvature, $\mu$ is the anisotropic curvature coefficient, $\overrightarrow{N}$ and $\overrightarrow{T}$ denote the unit inner normal and the unit tangent vector of $\gamma(u,t)$, respectively, and $s$ is the arclength parameter. The term $\gamma_1\overrightarrow{N}_0$ represents the initial velocity of the initial curve $\gamma_0$.
	
	In the level set method, the evolving curve $\gamma$ is represented by the zero level set of a function $\phi: \mathbb{R}^2\times[0,T)\rightarrow\mathbb{R}$, such that
	\begin{displaymath}
		C=\left\{x\in\mathbb{R}^2: \phi(x,t)=0\right\}.
	\end{displaymath}
	Since the flow \Cref{eq:31} is normal \cite{Deckelnick2023Discrete}, the evolution equation reduces to
	\begin{equation}\label{eq:33}
		\frac{\partial ^2 \gamma}{\partial t^2}(u,t)+\beta\frac{\partial\gamma}{\partial t}(u,t)=\mu\kappa(u,t).
	\end{equation}
	Let $v$ denote the normal velocity, \Cref{eq:33} can be reformulated as follows
	\begin{equation}\label{eq:34}
		v_t+\beta v=\mu\kappa.
	\end{equation}
	Differentiating the equality $\phi(x(t),t)=0$ with respect to $t$ yields
	\begin{displaymath}
		\frac{d}{dt}\phi=\phi_t+\nabla\phi\cdot v\overrightarrow{N}=0,
	\end{displaymath} 
	where $\overrightarrow{N}=\frac{\nabla\phi}{|\nabla\phi|}$. Then, the level set representation of the normal velocity is
	\begin{equation}\label{eq:35}
		v=-\frac{\phi_t}{|\nabla\phi|},
	\end{equation}
	and the normal acceleration is
	\begin{equation}\label{eq:36}
		v_t=\frac{d}{dt}\left(-\frac{\phi_t}{|\nabla\phi|}\right)
		=-\frac{1}{|\nabla \phi|}\left(\phi_{tt}
		+(\nabla\phi)_t\cdot v \cdot\frac{\nabla\phi}{|\nabla\phi|}\right)
		+\phi_t\cdot\frac{\nabla\phi\cdot(\nabla\phi)_t}{|\nabla\phi|^3}.
	\end{equation}
	As usual, we introduce the signed distance function to the interface $C$ as
	\begin{displaymath}
		d(x,t)=
		\begin{cases}
			\displaystyle \inf_{y\in C} \|x-y\|,\quad if\; x\in E_t,\\   	
			\displaystyle -\inf_{y\in C} \|x-y\|,\quad otherwise,
		\end{cases}
	\end{displaymath}
	where $E_t=\left\{x\in\mathbb{R}^2|\phi(x,t)>0\right\}$ and $C=\partial E_t$. Since the signed distance function satisfies $|\nabla d|=1$ i.e. $|\nabla\phi|=1$ (cf. \cite{li2005level}), we have
	\begin{displaymath}
		(\nabla\phi)_t\cdot\nabla\phi=0.
	\end{displaymath}
	Substituting this into \Cref{eq:36} gives rise to 
	\begin{equation}\label{eq:37}
		v_t=-\frac{\phi_{tt}}{|\nabla\phi|}.
	\end{equation}
	The curvature in level set form is
	\begin{equation}\label{eq:38}
		\kappa=\nabla\cdot\left(\frac{\nabla\phi}{|\nabla\phi|}\right).
	\end{equation}
	Using \Cref{eq:35}, \Cref{eq:37} and  \Cref{eq:38}, the model \Cref{eq:34} becomes
	\begin{equation}\label{eq:39}
		\phi_{tt}+\beta\phi_t=\mu|\nabla\phi|\nabla\cdot\left(\frac{\nabla\phi}{|\nabla\phi|}\right),
	\end{equation}
	and the initial conditions \Cref{eq:32} translate into
	\begin{equation}\label{eq:310}
		\phi(x,0) = \phi_0(x), \; \phi_t(x,0) = \phi_1(x),
	\end{equation}
	where $\phi_0$ is the level set formulation for initial curve $\gamma_0$, and $\phi_1$ is corresponds to initial normal velocity.
	
	Motivated by the effectiveness of the classical Chan-Vese model and the capability of deep priors to capture complex spatial structures, we adopt the hyperbolic mean curvature flow as a natural mathematical foundation to integrate these insights into a unified segmentation framework. Building upon the level set formulation derived above, in this paper we propose the following LHMCF model:
	\begin{equation}\label{eq:311}
		\left\{\begin{array}{ll}
			\phi_{tt}+\beta\phi_t=|\nabla\phi|\left(\mu\nabla\cdot\left(\frac{\nabla\phi}{|\nabla\phi|}\right)
			-\left(\lambda_1\|\mathbb{Z}-c_1\|_2^2-\lambda_2\|\mathbb{Z}-c_2\|_2^2\right)+\alpha R(\phi)\right), \\
			\phi(x,0) = \phi_0(x),  \\
			\phi_t(x,0) = \phi_1(x),
		\end{array}\right.
	\end{equation}
	where $\lambda_1, \lambda_2, \alpha>0$ are fixed parameters, $\mathbb{Z}\in\mathbb{R}^{H\times W\times D}$ denotes the high‑dimensional feature representation extracted by a backbone $\mathcal{F}$, and $c_1, c_2$ are the corresponding region mean in feature space. The term $R(\phi)$  represents a data‑driven deep prior that preserves the underlying semantic structures and provides additional regularization for the evolving level‑set function. Instead of estimating pixel‑wise differences in the intensity domain, the model evaluates region homogeneity in the high‑dimensional feature space, which strengthens the stability of the evolution and provides richer structure than traditional pixel‑level formulations. 
	
	Until now, we have developed a segmentation model that integrates curvature‑driven geometric evolution, feature space region statistics, and deep structural priors within a unified high‑dimensional framework.
	
	\subsection{Numerical formulation}
	To solve the proposed LHMCF model, we reformulate the second-order PDE in \Cref{eq:311} into a coupled system of two first-order equations
	\begin{align}
		v_t &= -\beta v + F(\phi(t)), \label{eq:312}\\
		\phi_t &= v, \label{eq:313}
	\end{align}
	where the synthesized driving force $F(\phi(t))$ is defined as
	\begin{displaymath}
		F(\phi(t))=|\nabla\phi|\left(\mu\nabla\cdot\left(\frac{\nabla\phi}{|\nabla\phi|}\right)
		-\left(\lambda_1\|\mathbb{Z}-c_1\|_2^2-\lambda_2\|\mathbb{Z}-c_2\|_2^2\right)+\alpha R(\phi)\right).
	\end{displaymath}
	\Cref{eq:312} can be viewed as a first-order non-homogeneous ordinary differential equation with respect to $v$. Multiplying both sides by  $e^{\beta t}$, the local exact solution over a time interva $[t_k, t_{k+1}]$ is given by Duhamel's principle,
	\begin{equation}\label{eq:314}
		v^{k+1}=v^k e^{-\beta\Delta t}+\int_{t_k}^{t_{k+1}}e^{-\beta(t_{k+1}-\tau)}F(\phi(\tau))d\tau,
	\end{equation}
	where $\Delta t=t_{k+1}-t_k$, $v^k=v(t_k)$, and $v^{k+1}=v(t_{k+1})$.
	For a sufficiently small time step $\Delta t$, the highly non-linear force $F(\phi(\tau))$ can be approximated as piecewise constant within the interval $[t_k, t_{k+1}]$. Specifically, we employ the left-endpoint rule by setting $F^k\equiv F(\phi(t_k)) \approx F(\phi(\tau))$ for $\tau \in [t_k, t_{k+1}]$. Under this locally linearized approximation, the integration of \Cref{eq:314} yields the velocity updatation
	\begin{equation}
		v^{k+1} = v^ke^{-\beta\Delta t} + F^k \int_{t_k}^{t_{k+1}} e^{-\beta(t_{k+1} - \tau)} d\tau = v^ke^{-\beta\Delta t} + \frac{F^k}{\beta}\left(1-e^{-\beta\Delta t}\right). \label{eq:315}
	\end{equation}
	
	Subsequently, the evolution of the level-set function $\phi$ is obtained by integrating \Cref{eq:313},
	\begin{displaymath}
		\phi^{k+1} = \phi^k + \int_{t_k}^{t_{k+1}} v(t) dt.
	\end{displaymath}
	By substituting \Cref{eq:315} into the above integral, we obtain the updating rule for $\phi$
	\begin{align}\label{eq:316}
		\phi^{k+1} &= \phi^k + \int_{t_k}^{t_{k+1}} \left[ v^ke^{-\beta (\tau-t_k)} + \frac{F^k}{\beta}\left(1 - e^{-\beta (\tau-t_k)}\right) \right] d\tau \\
		&=\phi^k+\frac{1-e^{-\beta\Delta t}}{\beta}v^k+\left(\Delta t-\frac{1-e^{-\beta\Delta t}}{\beta} \right)\frac{F^k}{\beta},\nonumber
	\end{align}
	where $\phi^k=\phi(t_k)$, and $\phi^{k+1}=\phi(t_{k+1})$.
	
	The equations \eqref{eq:315} and \eqref{eq:316} constitute an explicit exponential integration scheme for the coupled position-velocity system. This formulation provides the mathematical foundation for the deep unfolding implementation of the LHMCF-Net.
	
	As for $c_1$ and $c_2$, the feature means of the foreground and background regions, respectively, their estimates in the high-dimensional feature space $\mathbb{Z}$ can be expressed as
	\begin{equation}\label{eq:317}
		c_1^k = \frac{\int_{\Omega} \mathbb{Z}(x)\cdot H_{\epsilon}(\phi^k(x))dx}{\int_{\Omega}H_{\epsilon}(\phi^k(x))dx},\;
		c_2^k = \frac{\int_{\Omega} \mathbb{Z}(x)\cdot(1-H_{\epsilon}(\phi^k(x)))dx}{\int_{\Omega}(1-H_{\epsilon}(\phi^k(x)))dx},
	\end{equation}
	where $H_{\epsilon}$ is a smooth Heaviside function. \Cref{eq:317} can be interpreted as performing Masked Average Pooling (MAP) on the features, for both the foreground and background. While this operator provides a direct observation of regional features, it is sensitive to noise and boundary inaccuracies in a single stage. To establish a robust long-term information path and maintain consistency across the unfolding stages, these coarse estimates will be further refined through a momentum updating  approach introduced in the next section. This refinement strategy ensures $c_1$ and $c_2$ evolve steadily, providing reliable guidance for the hyperbolic flow.
	
	\subsection{Deep unfolding algorithm}
	\label{deep unfolding algorithm}
	To translate the continuous hyperbolic mean curvature flow into a computable feed-forward architecture, we adopt a deep unfolding strategy. This paradigm maps the iterative numerical procedure of the PDE into a sequence of discrete stages, where each stage corresponds to one evolution step of the underlying dynamical system. Unlike classical variational formulations with fixed priors, the unfolding framework enables the curvature term, the feature‑space data term, and the deep structural prior to be learned and refined end‑to‑end.
	
	The unfolded system begins with an initialization phase. A feature extraction backbone $\mathcal{F}$ maps the input image into a normalized high-dimensional feature space $\mathbb{Z}\in\mathbb{R}^{H\times W\times D}$. An initial level-set function $\phi^0$ is predicted by a lightweight convolution operator, from which the foreground and background feature means $(c_1^0, c_2^0)$ are computed using the MAP formulation in \Cref{eq:316}. The initial velocity $v^0$ is set to zero, corresponding to a stationary initial state. Since the learned parameters $\Theta=(\beta, \mu, \lambda_1, \lambda_2, \alpha, \Delta t)$ have strict physical meanings and mathematical constraints, we introduce a latent‑variable mapping to ensure physically valid initialization. The coefficients $\beta, \lambda_1, \lambda_2, \alpha$ are generated from zero‑initialized latent variables through the Softplus activation, guaranteeing positivity and yielding smooth initial values around $ln2$. To satisfy the Courant–Friedrichs–Lewy (CFL) stability condition, the time step is constrained by
	\begin{displaymath}
		\Delta t=\Delta t_{max}\cdot\sigma(\theta_{dt}),\quad
		\Delta t_{max}=0.35, 
	\end{displaymath}
	where $\sigma$ is the Sigmoid function and $\theta_{dt}$ is initialized to $-1.0$, producing a conservative initial step size. The anisotropic curvature coefficient $\mu$ is obtained by mapping $\mathbb{Z}$ through a $1\times1$ convolution followed by a Sigmoid activation function, allowing spatially adaptive curvature modulation.
	
	After initialization, the discrete evolution unfolds over $K$ sequential stages. At each stage $k$, the algorithm performs an alternating update scheme to evolve both the geometric state $(\phi^k, v^k)$ and the foreground/background feature means $(c_1^k, c_2^k)$. Coarse feature means are first computed via MAP using \Cref{eq:317}. To obtain a stable foreground representation, we adopt a two‑branch update mechanism. The stage‑wise memory of the foreground feature mean is updated by a momentum‑based operator
	\begin{displaymath}
		(c_1^k, c_1^{refined})=\mathrm{MFE}(c_1^{k-1}, c_1^{coarse}),
	\end{displaymath}
	where $c_1^{refined}$ denotes a refinement of the coarse observation produced by the  momentum-based feature evolution (MFE), and $c_1^k$ represents the updated stage‑wise memory.  MFE is a momentum‑based update operator that models the dependency between current observations and accumulated historical information. In parallel, the foreground feature mean used in  $F^k$ is obtained by a gated fusion
	\begin{displaymath}
		\tilde{c}_1^{k-1}=\mathrm{Norm}(g\cdot c_1^{refined}+(1-g)\cdot c_1^{coarse}),
	\end{displaymath}
	where $g$ is a learnable gate controlling the fusion between refined and coarse estimates. For the background feature mean, we employ an Exponential Moving Average (EMA)
	\begin{displaymath}
		c_2^k=\mathrm{Norm}(\eta c_2^{k-1}+(1-\eta)\cdot c_2^{coarse}),\quad\tilde{c}_2^{k-1}=c_2^k.
	\end{displaymath}
	where $\eta\in(0,1)$ is a momentum coefficient controlling the balance between historical information and the current coarse estimate. This updatation  introduces physical inertia and suppresses oscillations in the background feature field during the hyperbolic evolution.
	
	The deep prior $R(\phi^k)$ is generated by a Mask Denoiser (MD) module. Curvature is computed using \Cref{eq:38}, and the total force $F^k$ is synthesized from the curvature term, the feature‑space data term, and the learned prior,
	\begin{displaymath}
		F^{k-1} = |\nabla \phi^{k-1}|\left(\mu\nabla\cdot\left(\frac{\nabla\phi^{k-1}}{|\nabla\phi^{k-1}|}\right) 
		- \left(\lambda_1 \|\mathbb{Z}-\tilde{c}_1^{k-1}\|_2^2 - \lambda_2 \|\mathbb{Z}-\tilde{c}_2^{k-1}\|_2^2\right)
		+ \alpha R(\phi^{k-1})\right).
	\end{displaymath}
	The velocity and level‑set function are then updated using the discrete evolution rules 
	\begin{displaymath}
		v^k=v^{k-1}e^{-\beta\Delta t}+\frac{F^{k-1}}{\beta}\left(1-e^{-\beta\Delta t}\right),
	\end{displaymath}
	and
	\begin{displaymath}
		\phi^k =\phi^{k-1}+\frac{1-e^{-\beta\Delta t}}{\beta}v^{k-1}+\left(\Delta t-\frac{1-e^{-\beta\Delta t}}{\beta} \right)\frac{F^{k-1}}{\beta}.
	\end{displaymath}
	
	Taken together, the updates of the feature means $(c_1^k, c_2^k)$, the velocity field $v^k$, and the level‑set function $\phi^k$ constitute one complete unfolded evolution step of the hyperbolic flow. Each stage integrates feature‑space refinement, curvature‑driven geometry, and learned priors into a unified numerical update, thereby mirroring the behavior of the continuous dynamical system within a discrete, trainable architecture. The entire procedure is summarized in \cref{alg:unfolding}.
	
	\begin{algorithm}
		\caption{Deep Unfolding Hyperbolic Mean Curvature Flow}
		\label{alg:unfolding}
		\begin{algorithmic}		
			\STATE \textbf{Input:} image $x$, number of stages $K$.
			\STATE \textbf{Output:} level-set function $\phi^K$, segmentation mask $u^K$.		
			\STATE \textbf{Feature extraction:} $\mathbb{Z} \leftarrow \mathrm{Norm}(\mathcal{F}(x))$.
			\STATE \textbf{Initialization:}		
			\STATE {\centering
				\(
				v^0 \leftarrow 0, \;
				\phi^0 \leftarrow \mathrm{Conv}(\mathbb{Z}),\; c_1^0\leftarrow \mathrm{MAP}(\mathbb{Z},\; H(\phi^0)), c_2^0\leftarrow \mathrm{MAP}(\mathbb{Z}, 1-H(\phi^0)).
				\)
				\par}
			
			\FOR{$k = 1 : K$}	
			\STATE \textbf{Coarse feature means:}
			\STATE {\centering
				\(
				c_1^{coarse}\leftarrow \mathrm{MAP}(\mathbb{Z}, H(\phi^{k-1})),\;
				c_2^{coarse}\leftarrow \mathrm{MAP}(\mathbb{Z}, 1-H(\phi^{k-1})).
				\)
				\par}
			\STATE \textbf{Foreground feature evolution:}
			\STATE {\centering
				\(
				(c_1^k, c_1^{refined}) \leftarrow \mathrm{MFE}(c_1^{k-1}, c_1^{coarse}),
				\)
				\par}
			\STATE {\centering
				\(
				\tilde{c}_1^{k-1}\leftarrow\mathrm{Norm}(g\cdot c_1^{refined}+(1-g)\cdot c_1^{coarse}).
				\)
				\par}
			\STATE \textbf{Background feature evolution:}
			\STATE {\centering
				\(
				c_2^k\leftarrow\mathrm{Norm}(\eta c_2^{k-1}+(1-\eta)\cdot c_2^{coarse}),
				\)
				\par}
			\STATE {\centering
				\(
				\tilde{c}_2^{k-1}\leftarrow c_2^k.
				\)
				\par}
			\STATE \textbf{Deep prior and curvature:} 	
			\STATE {\centering
				\(
				R(\phi^{k-1}) \leftarrow \mathrm{MD}(\phi^{k-1}),\; \kappa^{k-1}\leftarrow\nabla\cdot\left(\frac{\nabla\phi^{k-1}}{|\nabla\phi^{k-1}|}\right).
				\)
				\par}
			\STATE \textbf{Total force:}
			\STATE {\centering
				\(
				F^{k-1} \leftarrow |\nabla \phi^{k-1}|\left(\mu\kappa^{k-1}
				- \left(\lambda_1 \|\mathbb{Z}-\tilde{c}_1^{k-1}\|_2^2 - \lambda_2 \|\mathbb{Z}-\tilde{c}_2^{k-1}\|_2^2\right)
				+ \alpha R(\phi^{k-1})\right).
				\)
				\par}
			\STATE \textbf{Hyperbolic evolution update:}
			\STATE {\centering
				\(
				v^k\leftarrow v^{k-1} e^{-\beta\Delta t} + \frac{F^{k-1}}{\beta}(1-e^{-\beta\Delta t}),
				\)
				\par}
			\STATE {\centering
				\(
				\phi^k \leftarrow \phi^{k-1} + \frac{1-e^{-\beta\Delta t}}{\beta}v^{k-1} 
				+ \left(\Delta t - \frac{1-e^{-\beta\Delta t}}{\beta}\right)\frac{F^{k-1}}{\beta}.
				\)
				\par}
			\STATE \textbf{Segmentation mask:} $u^k\leftarrow H(\phi^k)$		
			\ENDFOR		
			\RETURN $\phi^K,\; u^K$		
		\end{algorithmic}
	\end{algorithm}
	
	\subsection{Learned hyperbolic mean curvature flow network}
\begin{figure}[htbp]
		\centering
		\includegraphics[width=\textwidth]{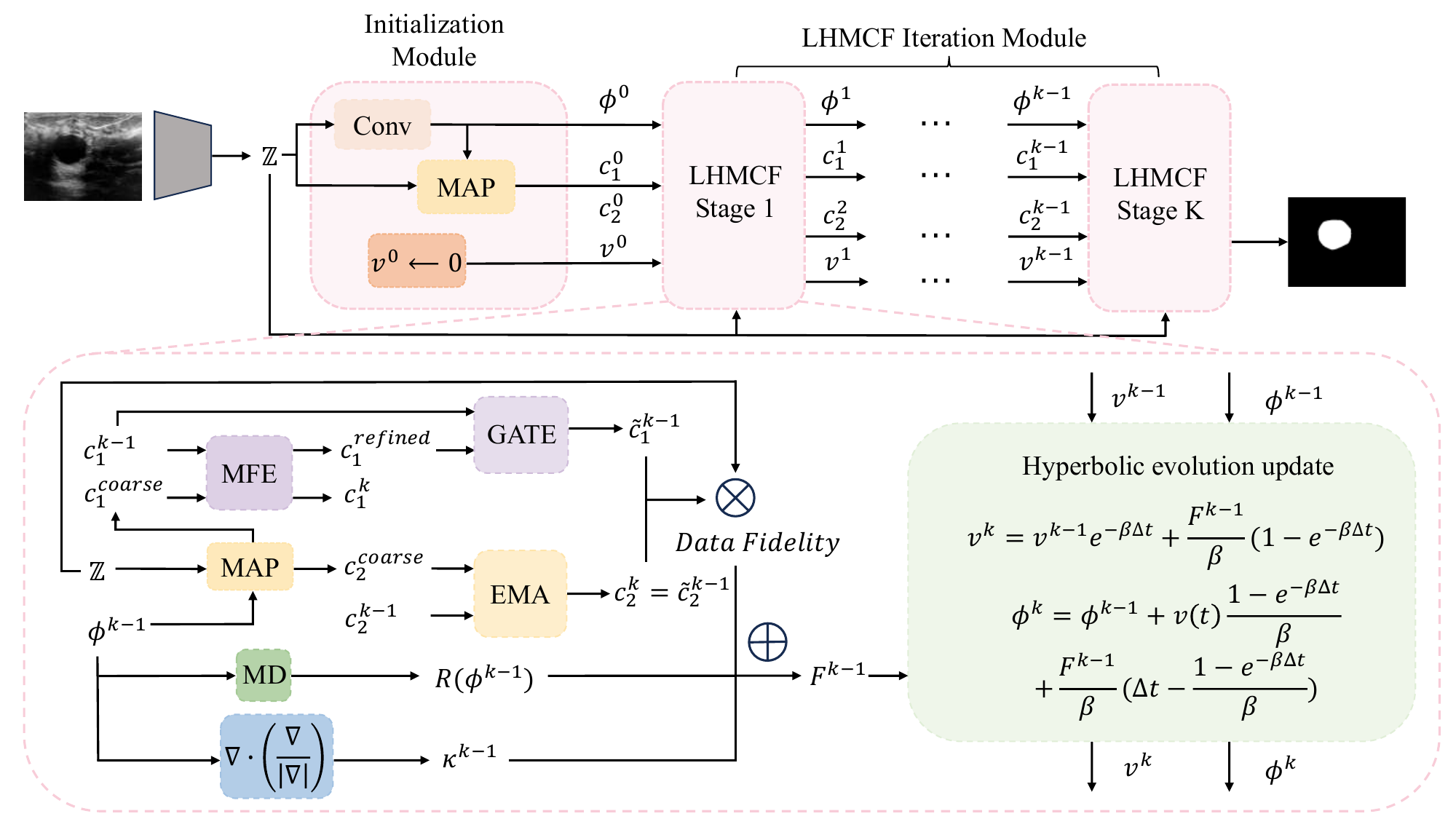}
		\caption{Overall structure of the proposed Learned Hyperbolic Mean Curvature Flow Network (LHMCF‑Net). The architecture comprises an Initialization Module that generates the initial level‑set function and velocity field, followed by the LHMCF Iteration Module, which performs hyperbolic evolution updates to iteratively refine the segmentation.}
		\label{fig:LHMCF}
	\end{figure}

	The discrete evolution in \cref{alg:unfolding} provides a stage-wise numerical procedure for solving the hyperbolic mean curvature model \Cref{eq:311}. Building upon this formulation, we construct a learned hyperbolic mean curvature network (LHMCF), illustrated in \Cref{fig:LHMCF}. The network consists of an Initialization Module and an LHMCF Iteration Module composed of $K$ unfolding stages.
	
	In the Initialization Module, we use a ResNet-101 backbone $\mathcal{F}$ as a feature extractor to map the input images $x$ to a high-dimensional feature space $\mathbb{Z}\in\mathbb{R}^{H\times W\times D}$. The initial level‑set function $\phi^0$ is generated by a $1\times1$ convolutional head acting on $\mathbb{Z}$, providing a data‑adaptive initialization consistent with the continuous formulation. Simultaneously, the initial velocity $v^0$ is set to zero. The initial foreground and background feature means $(c_1^0, c_2^0)$ are computed using the MAP operator in \Cref{eq:317}, providing statistically meaningful region descriptors for the first evolution stage. To enhance representational richness, the coarse foreground mask $H(\phi^0)$ is partitioned into $N$ uniform grid regions, and the MAP operator is applied independently to each region, yielding multiple representative foreground means through a self‑bootstrapping process.
	
	As shown in \Cref{fig:LHMCF}, the Iteration Module in LHMCF  consists of $K$ LHMCF stages, each corresponding to one evolution step of the discrete hyperbolic flow. Specifically, each stage includes a MAP operator, MFE, and EMA for updating the feature means, an MD module for refining the deep prior, as well as a hyperbolic evolution layer for implementing the damped PDE update.
	
\begin{figure}[htbp]
		\centering
		\includegraphics[width=0.7\textwidth]{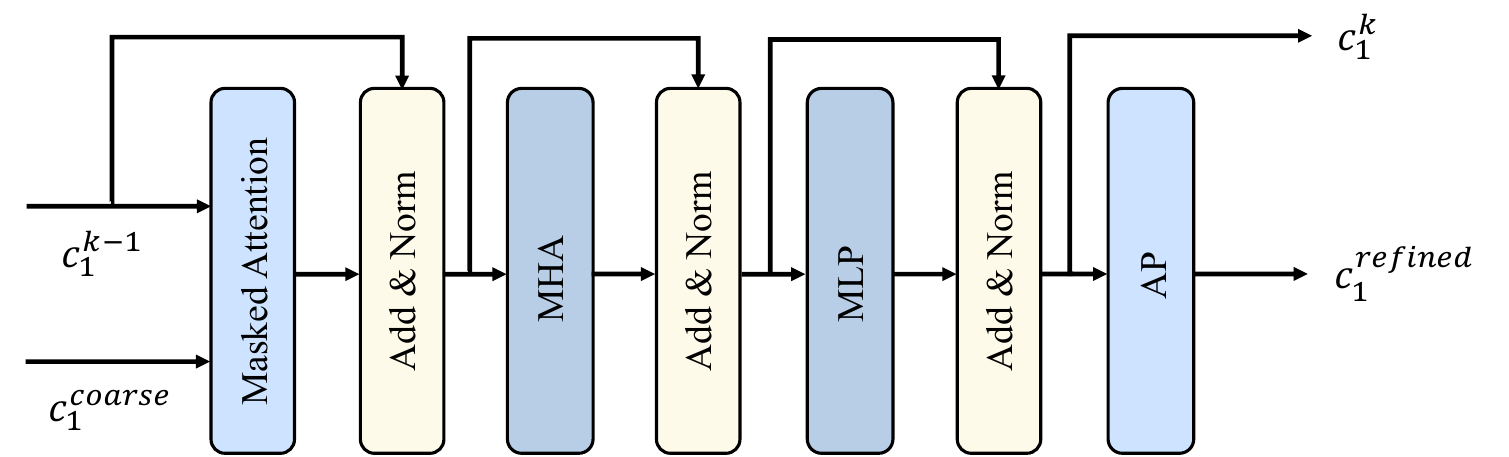}
		\caption{The structure of the proposed Momentum-based Feature Evolution.}
		\label{fig:MFE}
	\end{figure}

	The MFE module follows the MAP operator and performs a two‑branch update of the foreground feature representation, as illustrated in \Cref{fig:MFE}. Given the previous memory $c_1^{k-1}$ and the coarse foreground feature $c_1^{coarse}$, MFE produces both a memory update $c_1^k$, which stores stage‑wise semantic information, and a refined token $c_1^{refined}$, which is passed to the gated fusion mechanism to form the feature mean $\tilde{c}_1^k$ used in the data term of the force. This separation allows the memory branch to accumulate long‑range contextual information across unfolding stages, while the gated branch provides a stable and noise‑robust estimate for driving the hyperbolic evolution. The internal computation of MFE is defined as
	\begin{displaymath} 
		\begin{aligned} 
			&c^k = \mathrm{LN}(\mathrm{Softmax}(\mathcal{M}^k + c^{k-1} (c_1^{coarse})^T) c_1^{coarse} + c^{k-1}), \\ &c^k = \mathrm{LN}(\mathrm{MHA}(c^k, c^k, c^k) + c^k),\\
			&c^k = \mathrm{LN}(\mathrm{MLP}(c^k) + c^k), \\ &c_1^{refined} = \mathrm{AP}(c^k), 
		\end{aligned} 
	\end{displaymath}
	where $\mathrm{LN}$ denotes layer normalization, $\mathrm{MHA}$ denotes Multihead Self-Attention, $\mathrm{MLP}$ denotes Multi-Layer Perceptron and $\mathrm{AP}$ denotes average pooling. The masking matrix $\mathcal{M}\in\mathbb{R}^{N\times1}$ is defined by
	\begin{displaymath}
		\mathcal{M}^k(n)=
		\begin{cases}
			\displaystyle 0,\quad if\; c^{k-1}_n (c_1^{coarse})^T > \varepsilon^k,\vspace{2mm}\\   	
			\displaystyle -\infty,\quad otherwise,
		\end{cases} \quad n = 1, \dots, N,
	\end{displaymath}
	with the adaptive threshold
	\begin{displaymath}
		\varepsilon^k =\frac{1}{2}(min(c^{k-1}_n (c_1^{coarse})^T)+mean(c^{k-1}_n (c_1^{coarse})^T)).
	\end{displaymath}
	The masking mechanism ensures that only foreground tokens exhibiting sufficient similarity to the coarse feature participate in the update. The subsequent attention and MLP layers enhance the expressiveness of the refinement process, while average pooling aggregates the updated tokens into a representative foreground descriptor. Through this design, MFE enables reliable information propagation across all unfolding stages and mitigates the loss of long‑range dependencies.
	
\begin{figure}[htbp]
		\centering
		\includegraphics[width=0.7\textwidth]{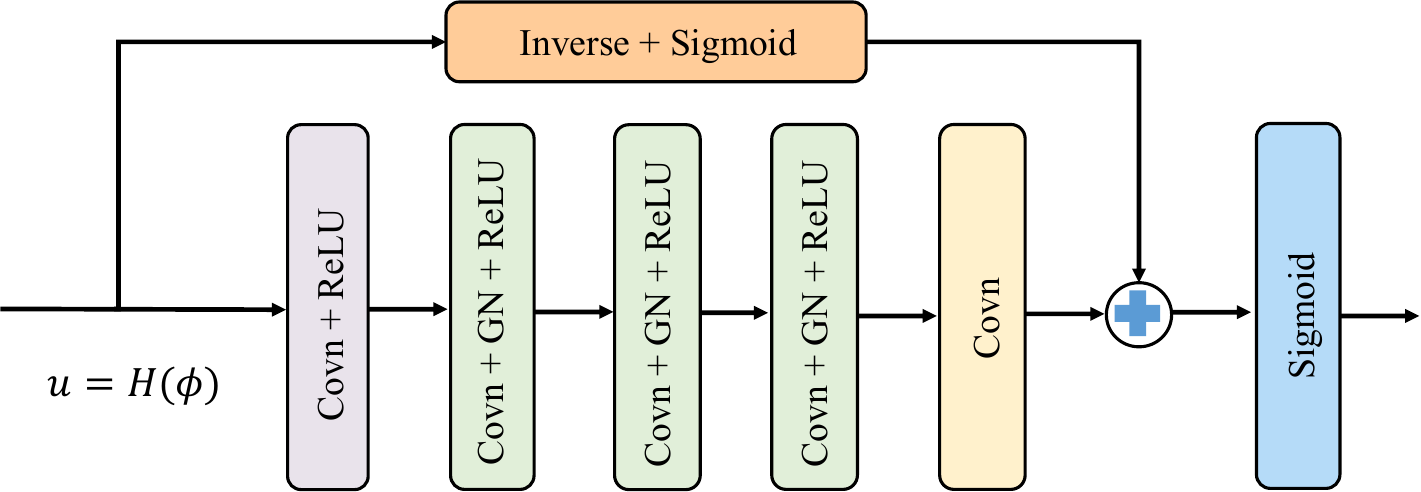}
		\caption{The structure of the proposed Mask Denoiser.}
		\label{fig:MD}
	\end{figure}

	The deep prior $R(\phi)$ is realized through an MD module, implemented as a 5-layer convolutional network with Group Normalization (GN) for stability under small batch sizes, illustrated in \Cref{fig:MD}. To avoid probability boundary saturation, the concatenated tensor is first mapped to an unbounded real-valued space using an inverse Sigmoid transformation. The MD predicts a geometric residual, which is subsequently projected back to the probability space, ensuring that the learned prior adheres to anatomical textures and complements curvature‑driven smoothing. Curvature $\kappa^k$ is computed using fixed convolutional stencils that discretize the geometric operators in the continuous model.
	
	The physical parameters $\Theta=(\beta, \mu, \lambda_1, \lambda_2, \alpha, \Delta t)$ are generated from unconstrained latent variables through monotone mappings (Softplus or Sigmoid), ensuring positivity and compliance with stability constraints such as the CFL condition. This parameterization maintains the operator structure of the underlying PDE while enabling the evolution dynamics to adjust to the statistical properties of the data. In this sense, deep unfolding provides a level of adaptivity that is not available in classical variational segmentation.
	
	The terminal component of each stage is the hyperbolic evolution layer, which implements the damped numerical scheme derived in \Cref{eq:315}-\Cref{eq:316}. This layer is deterministic and parameter‑free except for the learned physical parameters, preserving the exact discrete form of the hyperbolic flow. By embedding the PDE update directly into the network, LHMCF inherits the stability and interpretability of the continuous model while enabling end‑to‑end training.
	
	LHMCF-Net is not a conventional segmentation network but a learned dynamical system whose architecture mirrors the operator structure of the hyperbolic mean curvature flow.
	Each module corresponds to a specific operator in the discrete formulation, and each unfolding stage represents one evolution step of the underlying PDE. This tight coupling between mathematical modeling and neural design provides a principled framework for integrating geometric evolution, statistical region modeling, and learned priors, yielding a segmentation method that is both interpretable and data‑adaptive.
	
	\subsection{Loss function}
	To train the proposed LHMCF-Net in an end-to-end manner, we adopt a hybrid loss function that couples data-driven supervision with physics-informed geometric regularization. Let $P$ denote the probability prediction and $G$ represent the ground truth. The total loss $\mathcal{L}$ is defined as 
	\begin{equation}
		\mathcal{L} = \mathcal{L}_{Dice}(P, G) + \mathcal{L}_{BCE}(P, G) + \omega_{eik} \mathcal{L}_{Eikonal}(\phi) + \omega_{tv} \mathcal{L}_{TV}(\phi),
	\end{equation}
	where the weighting coefficients $\omega_{eik}=0.1$, $\omega_{tv}=0.01$ were determined empirically to balance the scale of gradients during backpropagation. The first two terms, Dice loss and Binary Cross-Entropy (BCE), provide strong semantic supervision by simultaneously maximizing region overlap and penalizing pixel-wise misclassification. The remaining two terms encode geometric priors on the evolving level-set function $\phi$. The Eikonal loss
	\begin{displaymath}
		\mathcal{L}_{Eikonal}=\frac{1}{|\Omega|}\int_\Omega(|\nabla\phi(x)|-1)^2dx,
	\end{displaymath}
	enforces the property of $|\nabla\phi|=1$ for the signed distance function, thereby eliminating the need for costly re-initialization procedures typically required in classical level-set methods. Finally, the Total Variation (TV) regularizer
	\begin{displaymath}
		\mathcal{L}_{TV}=\frac{1}{|\Omega|}\int_\Omega(|\partial_x\phi(x)|+|\partial_y\phi(x)|)dx
	\end{displaymath}
	acts as a global geometric smoothing prior that suppresses numerical oscillations and overshooting induced by second‑order hyperbolic dynamics, thereby promoting more stable evolution and improving boundary accuracy.
	
	During training, the loss is applied to all intermediate stages of LHMCF-Net, with stage-dependent weights gradually increasing from early to late iterations. This multi-stage supervision encourages consistent geometric refinement throughout the hyperbolic evolution process.
	
	\section{Experimental results}
	\label{sec:experiments}
	
	\subsection{Datasets}
	We evaluate the proposed method on three representative publicly available medical image datasets characterized by low contrast and ambiguous anatomical boundaries, including the following:
	
	\begin{itemize}
		\item Breast Ultrasound Images (BUSI) \cite{ALDHABYANI2020BUSI} consists of 780 ultrasound scans with an average image resolution of 500 $\times$ 500 pixels. Following common practice, we exclude the 133 normal scans and select the remaining 647 pathological scans (437 benign and 210 malignant tumors), each accompanied by a pixel‑level tumor mask.
	\end{itemize}
	
	\begin{itemize}
		\item Kvasir-SEG \cite{Jha2020KvasirSEG} contains 1,000 high-resolution gastrointestinal endoscopy images with polyp annotations. The image resolution ranges from 332 $\times$ 487 to 1920 $\times$ 1072 pixels, exhibiting substantial variation in scale, illumination, and boundary clarity.
	\end{itemize}
	
	\begin{itemize}
		\item ISIC 2018 \cite{Codella2018ISIC} provides 2,594 dermoscopic images released for the MICCAI 2018 Skin Lesion Analysis Challenge. The dataset includes diverse melanoma and non‑melanoma lesions with challenging visual characteristics such as irregular shapes, fuzzy borders, and strong color variability.
	\end{itemize}
	
	\subsection{Implementation details}
	Our framework is implemented in Python based on PyTorch and is trained end-to-end with a cascade stage $K$ set to 2. All experiments are conducted on an NVIDIA RTX A6000 GPU with 48 GB of memory. Following standardized evaluation protocols, the inputs for the BUSI and ISIC 2018 datasets are resized to 
	$256 \times 256$ pixels, while images from the Kvasir-SEG dataset are resized to $352 \times 352$ pixels. Our LHMCF-Net is trained episodically for 256 epochs using the AdamW optimizer with a batch size of 8. The initial learning rate is set to $1 \times 10^{-4}$ and a weight decay of $1 \times 10^{-4}$, accompanied by a Cosine Annealing learning rate scheduler. 
	
	\subsection{Evaluation metric}
	We will use Accuracy (Acc), Intersection over Union (IoU), Dice Similarity Coefficient (DSC), and 95th percentile Hausdorff Distance (HD95) \cite{huttenlocher1993comparing} to quantitatively evaluate segmentation performance. Let S represent the segmentation result and G the ground truth. In addition, TP, FP, FN and TN denote the numbers of true positives, false positives, false negatives, and true negatives, respectively. These metrics are defined as follows:
	
	Acc:
	\begin{displaymath}
		Acc(S, G) = \frac{TP + TN}{TP + FP + TN + FN}.
	\end{displaymath}
	
	IoU:
	\begin{displaymath}
		IoU(S, G) = \frac{|S \cap G|}{|S \cup G|} = \frac{TP}{TP + FP + FN}.
	\end{displaymath}
	
	DSC:
	\begin{displaymath}
		DSC(S, G) = \frac{2|S \cap G|}{|S| + |G|} = \frac{2TP}{2TP + FP + FN}.
	\end{displaymath}
	
	HD95:
	\begin{displaymath}
		HD95(S, G) = 
		\max \left\{
		P_{95}\!\left( \min_{y \in \partial G} \| x - y \| : x \in \partial S \right),
		P_{95}\!\left( \min_{x \in \partial S} \| y - x \| : y \in \partial G \right)
		\right\},
	\end{displaymath}
	where $\partial S$ and $\partial G$ denote the boundaries of the segmentation result and the ground truth, respectively, and $P_{95}$ denotes the 95th percentile.
	
	Among these metrics, Acc, IoU and DSC mainly emphasize the internal consistency of segmented objects, while the HD95 focuses on boundary precision. A lower HD95 value indicates more accurate boundary localization by the model.
	
	\subsection{Comparison with state-of-the-art methods}
	We compare our LHMCF-Net with several state-of-the-art methods, including UNet \cite{Ronneberger2015Unet}, UNet++ \cite{Zhou2020UNet++}, AttnUNet \cite{Oktay2018AttentionUNet}, Deeplab V3+ \cite{Chen2018DeepLabV3}, and TransUNet \cite{chen2021transunet}. The quantitative results on the BUSI, Kvasir, and ISIC 2018 datasets are summarized in \cref{tab:dice}. As shown, LHMCF-Net outperforms the other methods in terms of Acc, IoU, and DSC, while also achieving the lowest HD95. Specifically, LHMCF-Net reduces the HD95 by 1.4920, 0.8275 and 0.7955 compared with the second-best method on the BUSI, Kvasir and ISIC 2018 datasets, respectively, demonstrating its superior boundary localization capability.
	
	\begin{table}[htbp]
		\footnotesize
		\caption{Comparison of different segmentation methods on the BUSI, Kvasir and ISIC 2018 datasets. Best results are highlighted in bold.}\label{tab:dice}
		\begin{center}
			\begin{tabular}{l l l l l l l l} 
				\hline 
				Dataset & Metrics & UNet & UNet++ & AttnUNet & Deeplab V3+ & TransUNet & LHMCF-Net \\ 
				\hline 
				BUSI & Acc $\uparrow$ & 0.9616 & 0.9614 & 0.9708 & 0.9656 & 0.9720 & \textbf{0.9748}\\
				& IoU $\uparrow$ & 0.6525 & 0.6873 & 0.7569 & 0.7082 & 0.7649 & \textbf{0.7705}\\ 
				& DSC $\uparrow$ & 0.7747 & 0.8067 & 0.8562 & 0.8203 & 0.8613 & \textbf{0.8664} \\ 
				& HD95 $\downarrow$ & 26.4560 & 22.1092 & 18.4064 & 22.3252 & 16.1030 & \textbf{14.6110}\\ 
				\hline
				Kvasir & Acc $\uparrow$ & 0.9591 & 0.9660 & 0.9625 & 0.9705 & 0.9675 & \textbf{0.9718} \\
				& IoU $\uparrow$ & 0.7803 & 0.8136 & 0.7952 & 0.8338 & 0.8133 & \textbf{0.8440} \\ 
				& DSC $\uparrow$ & 0.8740 & 0.8959 & 0.8440 & 0.9081 & 0.9058 & \textbf{0.9136} \\ 
				& HD95 $\downarrow$ & 33.2971 & 31.0893 & 33.2521 & 27.0958 & 23.7453 & \textbf{22.9178} \\ 
				\hline
				ISIC 2018 & Acc $\uparrow$ & 0.9457 & 0.9477 & 0.9491 & 0.9495 & 0.9595 & \textbf{0.9624}  \\
				& IoU $\uparrow$ & 0.8058 & 0.8204 & 0.8227 & 0.8180 & 0.8312 & \textbf{0.8237} \\ 
				& DSC $\uparrow$ & 0.8898 & 0.8993 & 0.9002 & 0.8975 & \textbf{0.9069} & 0.9007 \\ 
				& HD95 $\downarrow$ & 16.2140 & 15.0142 & 15.0602 & 15.3753 & 12.9403 & \textbf{12.1448} \\ 
				\hline
			\end{tabular}
		\end{center}
	\end{table}
	
\begin{figure}[htbp]
		\centering
		\includegraphics[width=0.95\textwidth]{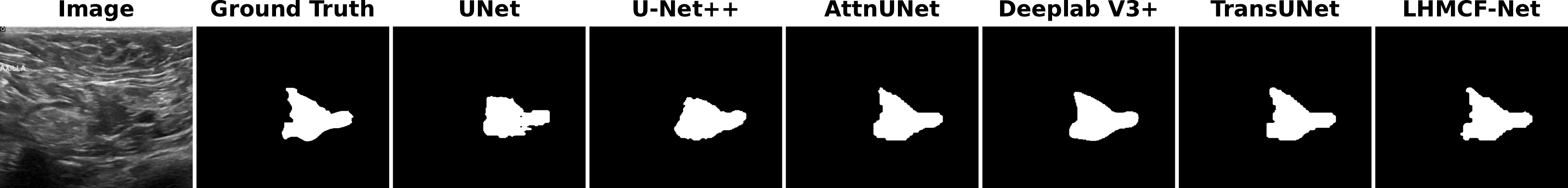} \\
		\vspace{0.1em}
		
		\includegraphics[width=0.95\textwidth]{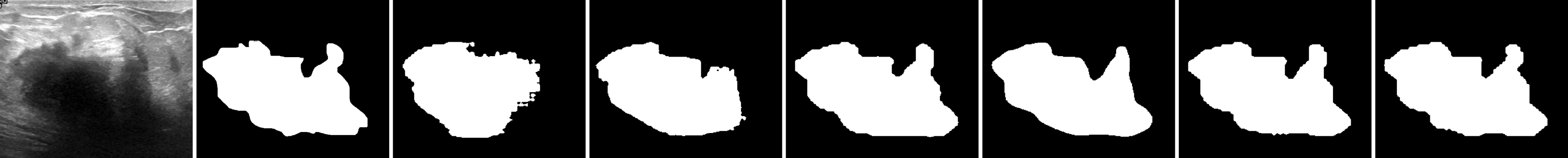} \\
		\vspace{0.1em}
		
		\includegraphics[width=0.95\textwidth]{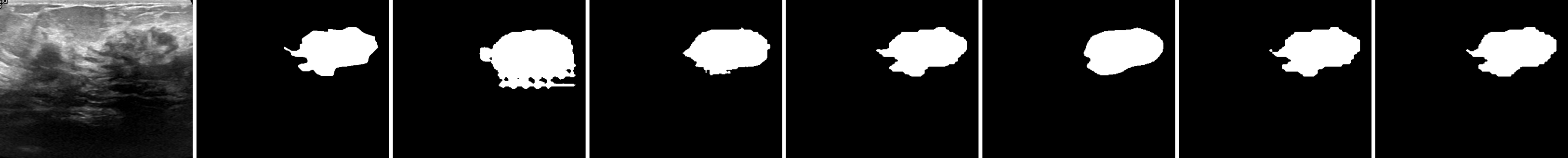} \\
		\vspace{0.1em}
		
		\includegraphics[width=0.95\textwidth]{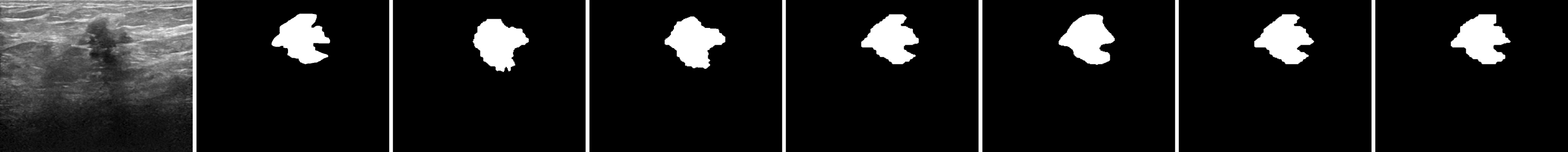} \\
		
		\caption{Qualitative comparison of our proposed LHMCF-Net with other methods on the BUSI dataset.}
		\label{fig:BUSI}
	\end{figure}

\begin{figure}[htbp]
		\centering
		\includegraphics[width=0.95\textwidth]{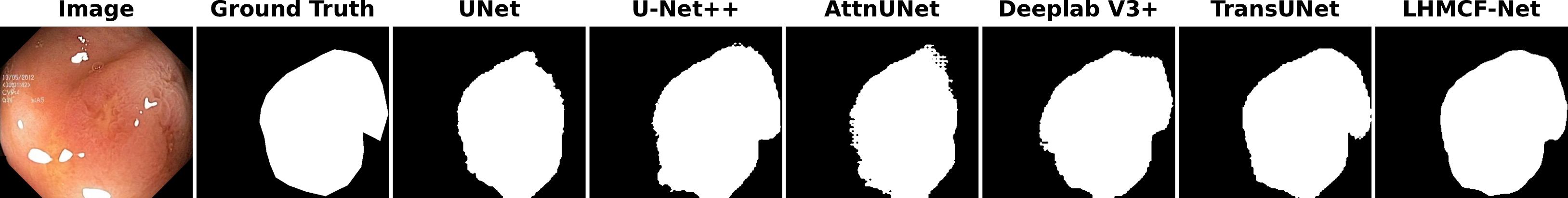}\\
		\vspace{0.1em}
		
		\includegraphics[width=0.95\textwidth]{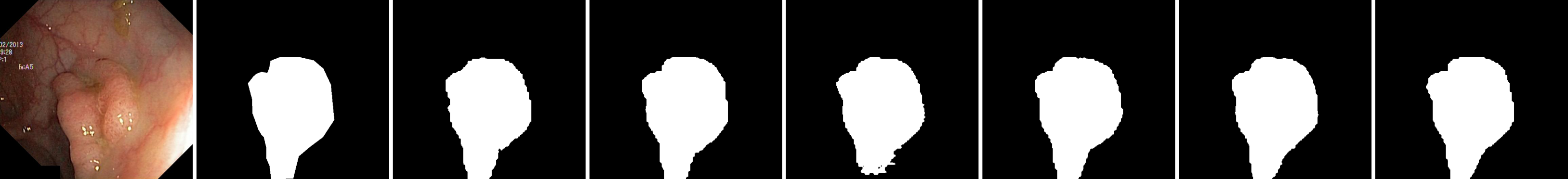}\\
		\vspace{0.1em}
		
		\includegraphics[width=0.95\textwidth]{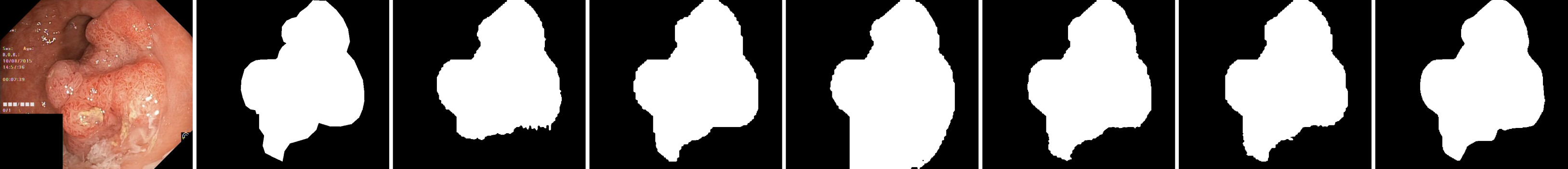}\\
		\vspace{0.1em}
		
		\includegraphics[width=0.95\textwidth]{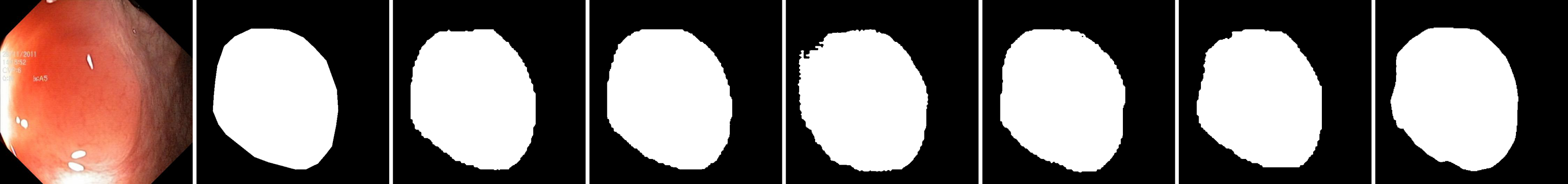}\\
		
		\caption{Qualitative comparison of our proposed LHMCF-Net with other methods on the Kvasir dataset.}
		\label{fig:Kvasir}
	\end{figure}

\begin{figure}[htbp]
		\centering
		\includegraphics[width=0.95\textwidth]{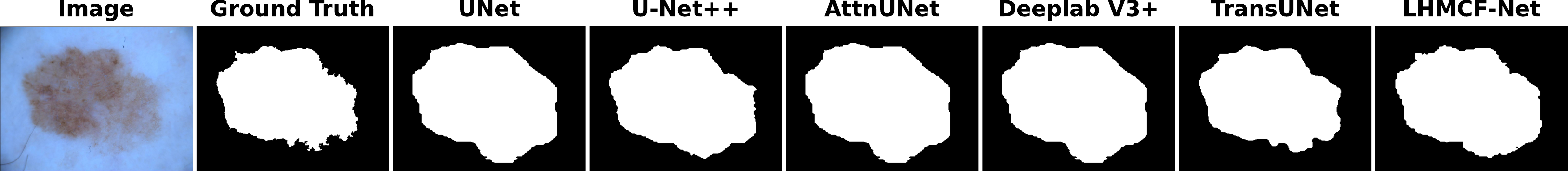} \\
		\vspace{0.1em}
		
		\includegraphics[width=0.95\textwidth]{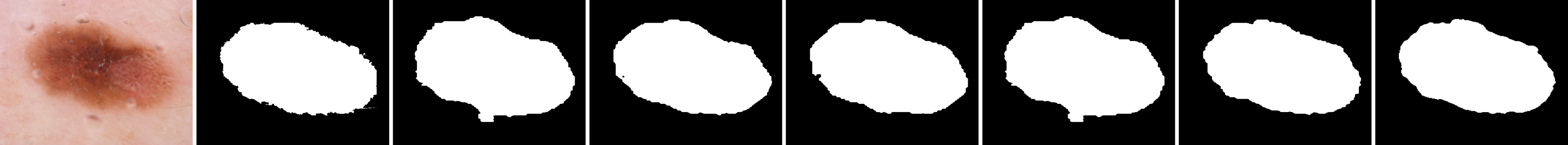} \\
		\vspace{0.1em}
		
		\includegraphics[width=0.95\textwidth]{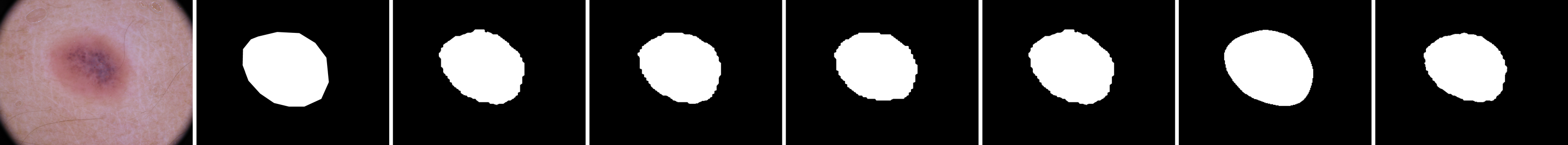} \\
		\vspace{0.1em}
		
		\includegraphics[width=0.95\textwidth]{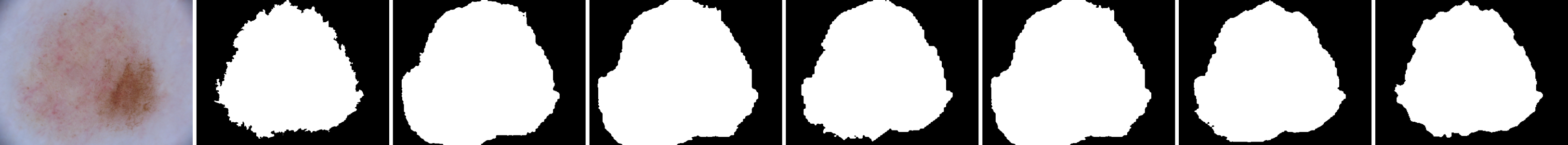} \\
		
		\caption{Qualitative comparison of our proposed LHMCF-Net with other methods on the ISIC 2018 datasets.}
		\label{fig:ISIC}
	\end{figure}

	To further illustrate the effectiveness of our approach, qualitative results are shown in \Cref{fig:BUSI}, \Cref{fig:Kvasir} and \Cref{fig:ISIC}.  LHMCF‑Net produces segmentation contours that adhere closely to the ground truth boundaries and exhibit smoother, more coherent shapes than those generated by other methods. This improvement arises from the integration of second‑order hyperbolic dynamics and curvature‑based geometric regularization, which jointly stabilize the interface evolution, suppress noise‑induced distortions, and prevent the contour from being trapped in shallow local minima. As a result, the model captures fine anatomical details while maintaining global topological consistency.
	
	\subsection[4.5]{Computational metrics comparison}
	\cref{tab:complexity} provides a comprehensive comparison of model complexity and computational efficiency across representative segmentation architectures. LHMCF‑Net achieves the smallest parameter count (36.69M), substantially lower than TransUNet (93.23M) and comparable U‑shaped CNNs, while maintaining Dice performance comparable to TransUNet and achieving the lowest HD95 values. This indicates that the physically guided hyperbolic formulation enables competitive accuracy without relying on large transformer backbones. Although LHMCF‑Net exhibits higher FLOPs due to the second‑order hyperbolic evolution and the feature‑space operators, its runtime remains competitive. The unfolding stages rely on lightweight convolutional updates that are highly parallelizable, mitigating the overhead introduced by the PDE dynamics. GPU memory usage is moderately higher because the model maintains additional physical fields (e.g., momentum and curvature coefficients), but the overall footprint remains within practical limits and comparable to transformer‑based methods. Overall, LHMCF‑Net offers a favorable trade‑off between model compactness and computational cost. The integration of interpretable physical dynamics with deep feature priors enables the network to match or surpass existing architectures while maintaining a significantly smaller parameter budget.
	
	\begin{table}[htbp]
		\footnotesize
		\caption{Comparison of parameters, runtime, FLOPs, and GPU memory of different methods. Best results are highlighted in bold.}\label{tab:complexity}
		\begin{center}
			\begin{tabular}{l l l l l l l} 
				\hline 
				Networks & UNet & UNet++ & AttnUNet & Deeplab V3+ & TransUNet & LHMCF-Net \\ 
				\hline 
				no. of parameters $\downarrow$ & 51.51 M & 67.98 M & 37.88 M & 45.67 M & 93.23 M & \textbf{36.69 M}\\
				\hline 
				FLOPs $\downarrow$ & 31.16 G & 124.87 G & 133.58 G & \textbf{28.20 G} & 64.46 G & 234.31 G\\ 
				\hline 
				test runtime $\downarrow$ & 30.64 ms & 34.21 ms & \textbf{9.98 ms} & 31.35 ms & 34.97 s & 42.23 s \\ 
				\hline
				train runtime $\downarrow$ & 100.84 ms & 111.45 ms  & \textbf{33.58 s} & 95.85 s & 148.47 s & 135.32 s\\
				\hline
				test memory $\downarrow$ & \textbf{342 M} & 810 M & 859 M & 339 M & 738 M & 1211 M \\
				\hline 
				train memory $\downarrow$ & 1194 M & 1978 M & 1348 M & \textbf{1054 M} & 2641 M & 2472 M  \\ 
				\hline
			\end{tabular}
		\end{center}
	\end{table}
	
	\section{Ablation analysis and physical interpretability}
	\label{sec:ablation}
	
	\subsection{Ablation on Hyperbolic vs. Parabolic Dynamics}
\begin{figure}[htbp]
		\centering
		\includegraphics[width=0.8\textwidth]{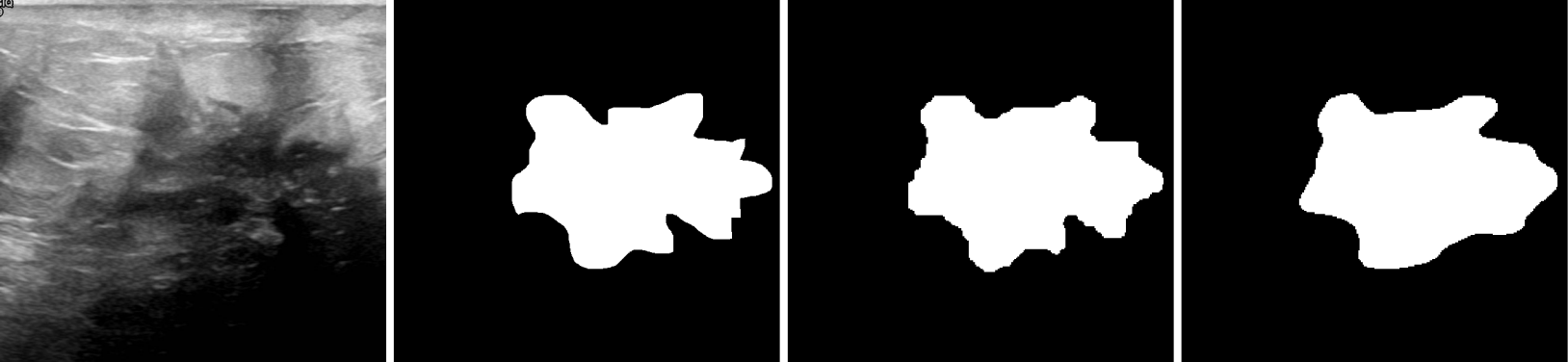} \\
		\vspace{0.8em}
		
		\includegraphics[width=0.8\textwidth]{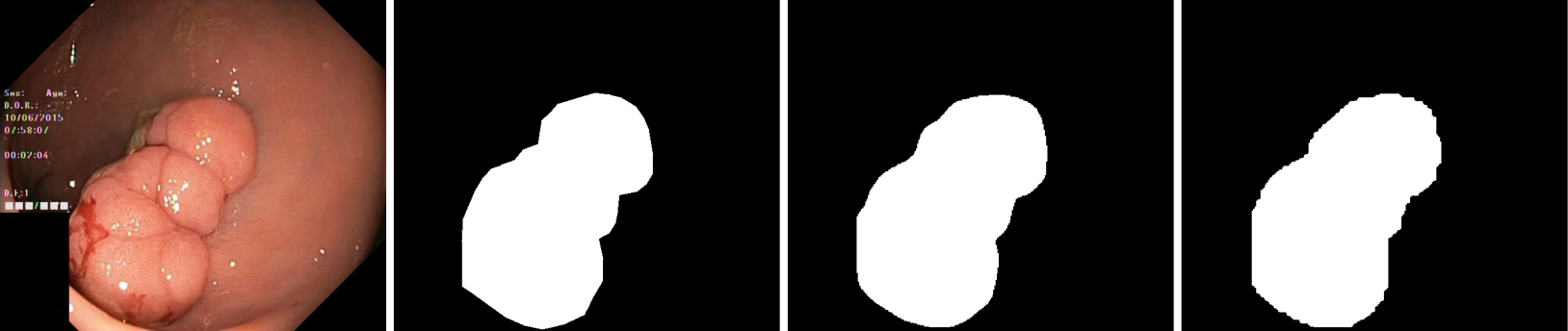} \\
		
		\caption{Qualitative comparison between hyperbolic and parabolic dynamics on BUSI and Kvasir datasets (from top to bottom). Each row displays original image, ground truth, LHMCF-Net, and LMCF-Net.}
		\label{fig:ablation_hyperbolic}
	\end{figure}

	\begin{table}[htbp]
		\footnotesize
		\caption{Comparison of segmentation performance between the proposed second-order hyperbolic model (LHMCF-Net) and the first-order parabolic counterpart (LMCF-Net) on the BUSI and Kvasir datasets. Best results are highlighted in bold.}\label{tab:hyper or para}
		\begin{center}
			\begin{tabular}{l l l l l l} 
				\hline 
				Dataset & Method & Acc $\uparrow$ & IoU $\uparrow$ & DSC $\uparrow$ & HD95 $\downarrow$ \\ 
				\hline 
				BUSI & LHMCF-Net & \textbf{0.9748} & \textbf{0.7705} & \textbf{0.8664} & \textbf{14.6110} \\
				& LMCF-Net & 0.9706 & 0.7518 & 0.8536 & 15.3634\\ 
				\hline
				Kvasir & LHMCF-Net & \textbf{0.9718} & \textbf{0.8440} & \textbf{0.9136} & \textbf{22.9178} \\
				& LMCF-Net & 0.9715 & 0.8382 & 0.9108 & 26.3625 \\ 
				\hline
			\end{tabular}
		\end{center}
	\end{table}
	To assess the contribution of the second‑order  term $\phi_{tt}$ to the overall segmentation performance, we performed an ablation study in which our model was degenerated into a first‑order parabolic formulation, denoted as LMCF‑Net. This reduction was implemented by removing the inertial term and fixing the damping coefficient to $\beta=1$, yielding a classical gradient descent type evolution. As demonstrated in \cref{tab:hyper or para}, the proposed LHMCF‑Net achieves consistently superior performance across both BUSI and Kvasir datasets. In particular, the HD95 metric decreases from 15.3634 to 14.6110 on BUSI and from 26.36 to 22.92 on Kvasir, indicating a clear advantage of the hyperbolic formulation. This improvement suggests that the introduction of physical inertia helps the evolving interface to bypass noise-induced local minima and better align with intricate anatomical boundaries. In contrast, the parabolic counterpart exhibits a more diffusive behavior, which tends to oversmooth fine structures and produces larger boundary displacement errors. These results validate that the hyperbolic evolution provides a more powerful and flexible trajectory for medical image segmentation than traditional first-order gradient descent flows. Qualitative comparisons are provided in \Cref{fig:ablation_hyperbolic}, further illustrating the visual differences between hyperbolic and parabolic evolutions.
	
	\subsection{Ablation study on damping coefficient}
	As illustrated in \cref{tab:beta}, the damping coefficient $\beta$ plays a critical role in shaping the hyperbolic evolution of the level‑set interface. We compare our autonomously learned $\beta$ against an undamped configuration $(\beta=0)$ and several fixed empirical settings spanning under-damped $(\beta=0.1)$ to strictly over-damped regimes $(\beta=10)$. When $\beta=0$, the model reduces to a purely hyperbolic system without any dissipative mechanism. This configuration exhibits noticeable instability in boundary evolution, reflected by higher HD95 values (15.5886 on BUSI and 25.5443 on Kvasir). The absence of damping allows oscillatory propagation of the interface, making it more susceptible to noise and causing the contour to overshoot or fluctuate around anatomical boundaries. For small damping values, the system becomes mildly under‑damped. Although this suppresses part of the oscillation, the improvement remains less obvious. Conversely, large damping values push the system into a strongly over‑damped regime. While this configuration stabilizes the evolution, it excessively suppresses the physical inertia needed to traverse low‑contrast regions or bypass shallow local minima. As a result, HD95 increases to 16.4565 and the DSC score drops to 0.8582 on BUSI, and the HD95 value rises to 26.5441 on Kvasir. These results confirm that over‑damping leads to overly diffusive behavior and boundary under segmentation.
	
	\begin{table}[htbp]
		\footnotesize
		\caption{Ablation study on the damping coefficient $\beta$ for the LHMCF model. Performance is evaluated under fixed damping configurations and the proposed learned damping strategy on the BUSI and Kvasir datasets. Best results are highlighted in bold.}\label{tab:beta}
		\begin{center}
			\begin{tabular}{l l l l l l} 
				\hline 
				Dataset & Damping Configuration & Acc $\uparrow$ & IoU $\uparrow$ & DSC $\uparrow$ & HD95 $\downarrow$ \\ 
				\hline 
				& $\beta=0$ & 0.9696 & 0.7572 & 0.8569 & 15.5886\\
				& $\beta=0.1$ & 0.9720 & 0.7568 & 0.8568 & 15.4733\\
				BUSI & $\beta=1$ & 0.9718 & 0.7574 & 0.8567 & 15.7718\\
				& $\beta=10$ & 0.9724 & 0.7587 & 0.8582 & 16.4565\\
				& Learned $\beta$(Ours) & \textbf{0.9748} & \textbf{0.7705} & \textbf{0.8664} & \textbf{14.6110} \\ 
				\hline
				& $\beta=0$ & 0.9701 & 0.8366 & 0.9094 & 25.5443 \\
				& $\beta=0.1$ & 0.9698 & 0.8307 & 0.9060 & 25.1687\\
				Kvasir & $\beta=1$ & 0.9678 & 0.8339 & 0.9078 & 25.6594\\
				& $\beta=10$ & 0.9713 & 0.8347 & 0.9081 & 26.5441\\
				& Learned $\beta$(Ours) & \textbf{0.9718} & \textbf{0.8440} & \textbf{0.9136} & \textbf{22.9178} \\		 
				\hline
			\end{tabular}
		\end{center}
	\end{table}
	
	By treating $\beta$ as a learnable parameter within our deep unfolding framework, the model adaptively balances inertia and dissipation according to the underlying anatomical structures. The learned $\beta$ consistently achieves the best performance across all metrics, validating that the learned dissipative dynamics provide an optimal trade‑off between stability and responsiveness, enabling more coherent interface propagation and significantly enhancing segmentation precision.
	
	\subsection{Ablation on deep prior term}
	In this ablation study, we remove the deep prior term $\alpha R(\phi)$ from \Cref{eq:311}, such that the hyperbolic evolution is driven solely by the mean curvature and the feature space data fidelity. As shown in \cref{tab:deep prior}, incorporating the deep prior consistently improves segmentation performance across both datasets. On BUSI, the DSC increases from 0.8513 to 0.8664 and HD95 decreases from 17.9611 to 14.6110. On Kvasir, the DSC improves from 0.9089 to 0.9136 and HD95 decreases from 26.9117 to 22.9178. These results indicate clear gains in both region accuracy and boundary fidelity. From a mathematical perspective, the feature‑space data term encourages regional homogeneity in $\mathbb{Z}$, but it lacks explicit geometric or semantic guidance. The deep prior $R(\phi)$, learned from high‑level representations, provides an implicit geometric regularization that steers the hyperbolic flow toward a plausible semantic manifold. This learned correction mechanism provides high-level semantic guidance that stabilizes the evolution and prevents the interface from drifting toward anatomically unrealistic configurations. Consequently, the final segmentation better preserves global anatomical structure while capturing fine‑scale boundary details.
	
	\begin{table}[htbp]
		\footnotesize
		\caption{Ablation study on the deep prior term in the LHMCF model. Performance is evaluated with and without the deep prior across the BUSI and Kvasir datasets. Best results are highlighted in bold.}\label{tab:deep prior}
		\begin{center}
			\begin{tabular}{l l l l l l} 
				\hline 
				Dataset & Method & Acc $\uparrow$ & IoU $\uparrow$ & DSC $\uparrow$ & HD95 $\downarrow$ \\ 
				\hline 
				BUSI & w/o deep proir & 0.9699 & 0.7485 & 0.8513 & 17.9611 \\
				& w deep prior & \textbf{0.9748} & \textbf{0.7705} & \textbf{0.8664} & \textbf{14.6110} \\	 
				\hline
				Kvasir & w/o deep prior & 0.9708 & 0.8356 & 0.9089 & 26.9117 \\
				& w deep proir & \textbf{0.9718} & \textbf{0.8440} & \textbf{0.9136} & \textbf{22.9178} \\ 
				\hline
			\end{tabular}
		\end{center}
	\end{table}
	
	\subsection{Ablation on Feature‑Mean Evolution}
	As introduced in \cref{deep unfolding algorithm}, the foreground and background feature means are dynamically evolved through a momentum‑based feature evolution operator MFE and EMA. To assess its contribution, we evaluate a baseline configuration in which the feature means used to compute the total force are obtained directly from the instantaneous coarse estimates through \Cref{eq:317}, without any temporal evolution. This results in a memory‑less update scheme where the feature means rely solely on potentially noisy observations from the current interface. As shown in \cref{tab:MFE}, this leads to degraded boundary accuracy, with HD95 increasing from 14.6110 to 17.1803 on BUSI and from 22.9178 to 26.7672 on Kvasir. The DSC also drops from 0.8664 to 0.8596 on BUSI and from 0.9136 to 0.9061 on Kvasir, indicating reduced regional consistency. In contrast, the full model incorporates historical inertia and global self‑attentive refinement, enabling the feature means to evolve smoothly along the semantic manifold of the target anatomy. This dynamic memory stabilizes the regional consistency term in the hyperbolic evolution \Cref{eq:311}, suppresses noise‑induced fluctuations, and prevents the interface from drifting toward anatomically implausible configurations. 
	
	\begin{table}[htbp]
		\footnotesize
		\caption{Ablation study on the momentum‑based feature evolution (MFE). Performance is evaluated with and without MFE on the BUSI and Kvasir datasets. Best results are highlighted in bold.}\label{tab:MFE}
		\begin{center}
			\begin{tabular}{l l l l l l} 
				\hline 
				Dataset & Method & Acc $\uparrow$ & IoU $\uparrow$ & DSC $\uparrow$ & HD95 $\downarrow$ \\ 
				\hline 
				BUSI & w/o MFE & 0.9721 & 0.7611 & 0.8596 & 17.1803 \\
				& w MFE & \textbf{0.9748} & \textbf{0.7705} & \textbf{0.8664} & \textbf{14.6110} \\ 
				\hline
				Kvasir & w/o MFE & 0.9689 & 0.8310 & 0.9061 & 26.7672  \\
				& w MFE & \textbf{0.9718} & \textbf{0.8440} & \textbf{0.9136} & \textbf{22.9178} \\ 
				\hline
			\end{tabular}
		\end{center}
	\end{table}
	
	\subsection{Impact of the number of stages $K$} 
\begin{figure}[htbp]
		\centering
		\begin{minipage}{0.32\textwidth}
			\centering
			\includegraphics[width=\textwidth]{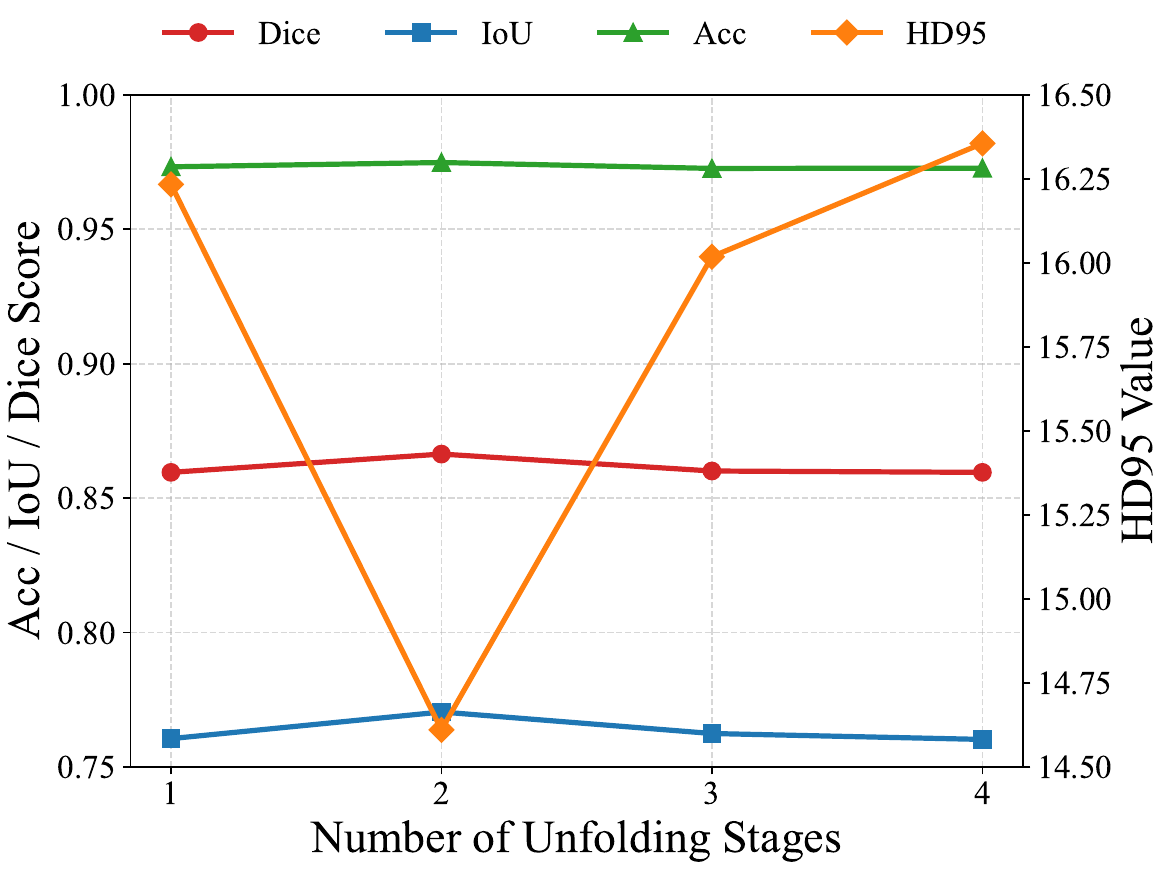}
		\end{minipage}\hfill
		\begin{minipage}{0.32\textwidth}
			\centering
			\includegraphics[width=\textwidth]{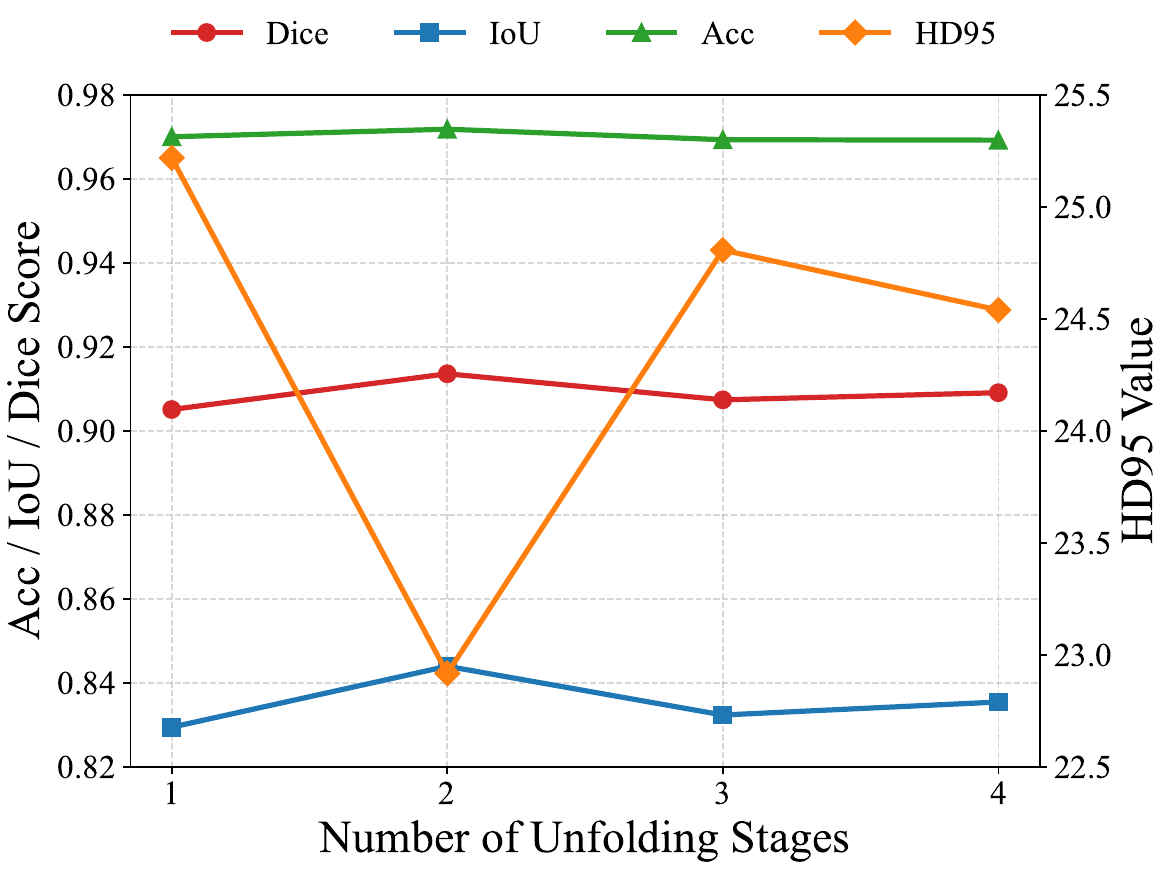}
		\end{minipage}\hfill
		\begin{minipage}{0.32\textwidth}
			\centering
			\includegraphics[width=\textwidth]{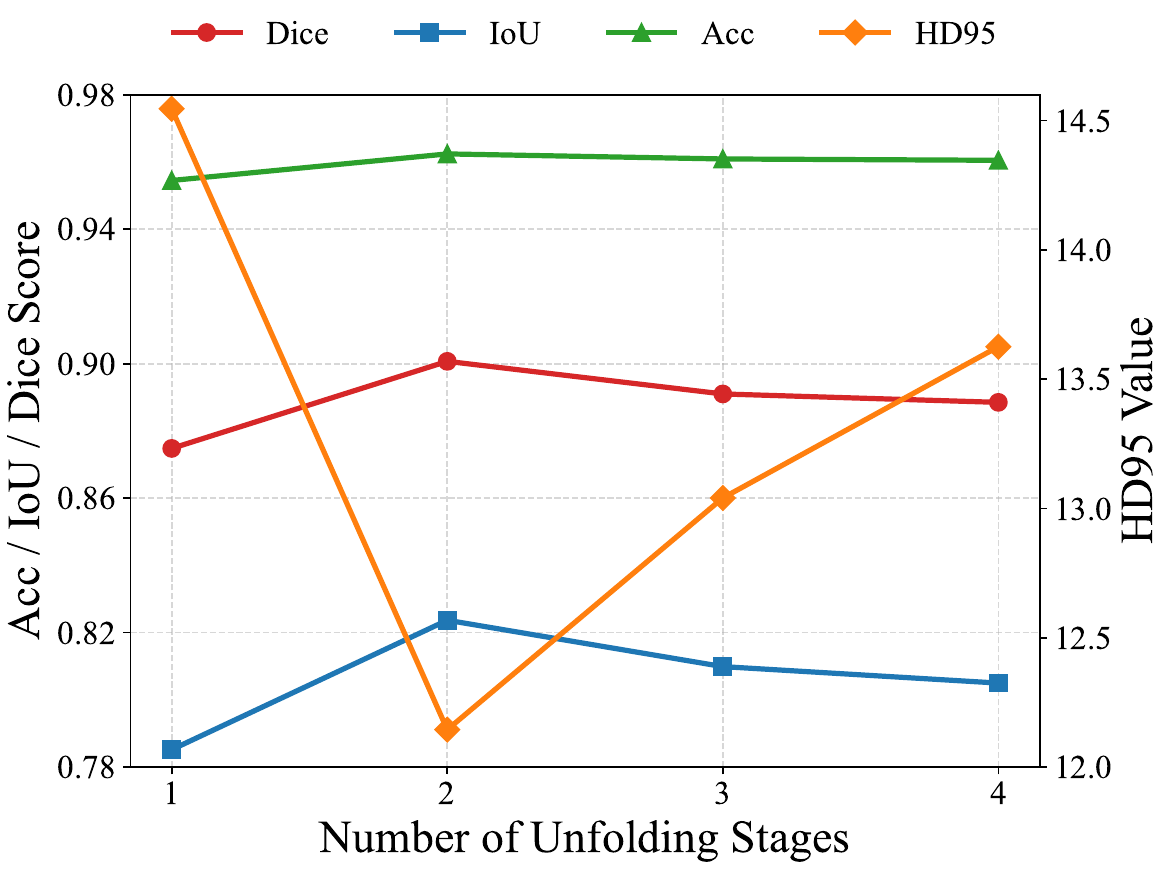}
		\end{minipage}
		\caption{Ablation on the number of unfolding stages ($K$). Results are shown for the BUSI, Kvasir, and ISIC 2018 datasets (from left to right).}
		\label{fig:stages_ablation}
	\end{figure}

	To investigate the influence of the unfolding depth, we conducted an ablation study by varying the number of stages  $K\in\{1, 2, 3, 4\}$. In our LHMCF framework, each stage corresponds to one discrete integration step of the second‑order hyperbolic system \Cref{eq:315}-\Cref{eq:316}. As illustrated in \Cref{fig:stages_ablation}, LHMCF-Net achieves its best performance at $K=2$ across the BUSI, Kvasir, and ISIC 2018 datasets. A single unfolding stage $K=1$ provides only a minimal amount of geometric evolution,  making it difficult for the level-set function $\phi$ to propagate through low‑contrast regions or to accurately align with complex anatomical boundaries. When the number of stages is further increased, say $K=3,4$, the performance no longer improves. Instead, deeper unrolling tends to introduce excessive smoothing, inertial overshooting, and numerical error accumulation. Overall, $K=2$ offers the most effective balance between representational capacity and numerical stability for the LHMCF‑Net.
	
	\subsection{Interpretability of learned parameters}
\begin{figure}[htbp]
		\centering
		\includegraphics[width=0.8\textwidth]{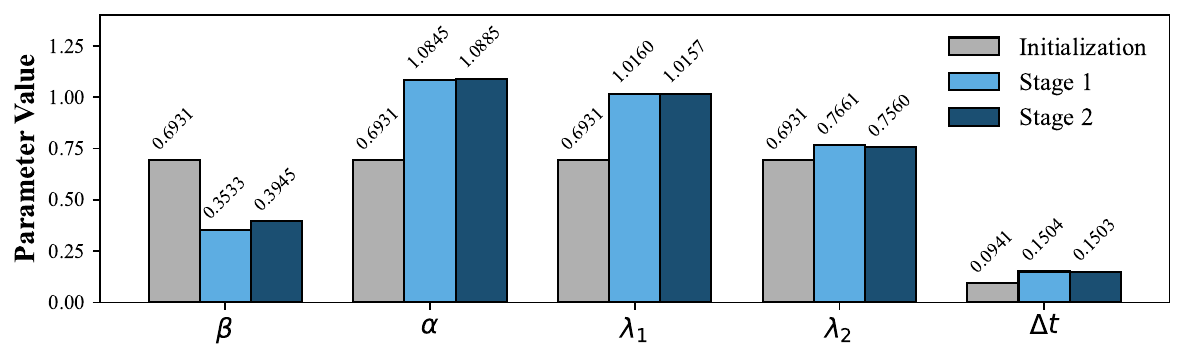} \\
		
		\includegraphics[width=0.8\textwidth]{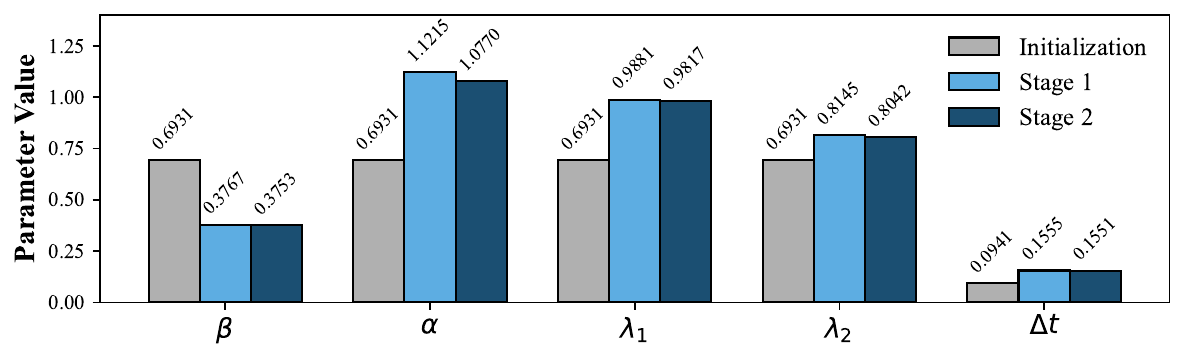}\\
		
		\includegraphics[width=0.8\textwidth]{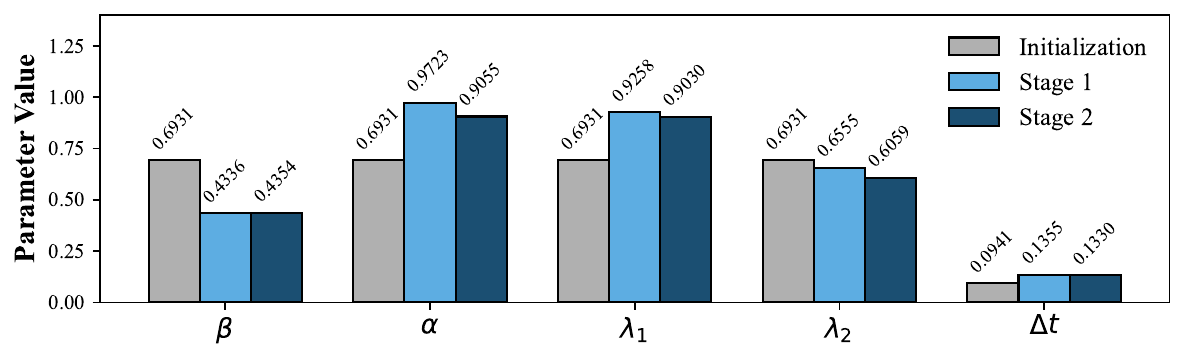}
		\\
		
		\caption{Evolution of the learned parameters $\beta,  \lambda_1, \lambda_2, \alpha, \Delta t$ across initialization and unfolding stages on the BUSI, Kvasir, and ISIC 2018 datasets (from top to bottom).}
		\label{fig:learned_parameters}
	\end{figure}

	To better understand the internal physical dynamics encoded by LHMCF, we analyze the autonomously learned parameters
	\begin{displaymath}
		\Theta=(\beta, \mu, \lambda_1, \lambda_2, \alpha, \Delta t)
	\end{displaymath}
	across unfolding stages, as shown in \cref{fig:learned_parameters}. These parameters are initialized through monotone mappings (e.g., Softplus, Sigmoid) to ensure physical validity and are optimized end‑to‑end under the hybrid segmentation loss.
	
	In the second-order hyperbolic evolution \Cref{eq:311}, the damping coefficient $\beta$ controls the rate of energy dissipation, while $\Delta t$ determines the integration step size of the exponential update scheme. Across all datasets, the learned $\beta$ decreases steadily from its initial value of 0.6931 to approximately 0.394 (BUSI), 0.353 (Kvasir), and 0.347 (ISIC 2018). In contrast,  $\Delta t$ increases from 0.094 to roughly 0.150–0.162. A reduced $\beta$ corresponds to weaker dissipation and thus stronger physical inertia. This indicates that the network favors a moderately under‑damped regime, allowing the interface to maintain sufficient momentum to traverse low‑contrast regions without being overly influenced by noise. The concurrent increase in $\Delta t$ accelerates the propagation of the interface, enabling the system to reach a stable semantic configuration within fewer unfolding stages. Together, these trends reveal that the learned hyperbolic dynamics balance responsiveness and stability more effectively than any fixed parameter choice.
	
	The parameter $\alpha$ regulates the influence of the deep prior $R(\phi)$.  We observe a significant upward trend, with $\alpha$ reaching 1.084 (BUSI), 1.164 (Kvasir), and 0.9055 (ISIC 2018). This increase suggests that, as the training progresses, the model places greater emphasis on geometric and semantic regularization rather than relying solely on regional homogeneity in the feature space $\mathbb{Z}$. Medical images often contain complex structures, ambiguous boundaries, and topology variations. The strengthened deep prior helps guide the hyperbolic flow toward anatomically plausible shapes, improving boundary coherence and reducing sensitivity to noise.
	
	The coefficients $\lambda_1$ and $\lambda_2$ weight the foreground and background homogeneity terms, respectively.
	Across all unfolding stages, the learned parameters consistently satisfy $\lambda_1>\lambda_2$. This asymmetric configuration reflects a deliberate emphasis on the foreground manifold. A larger $\lambda_1$ enforces stricter consistency between the evolving interface and the representative foreground feature mean $c_1$, ensuring that the segmentation remains tightly aligned with the target anatomy. Meanwhile, a smaller $\lambda_2$ provides tolerance for background variability, preventing the interface from being attracted to irrelevant structures or noise.
	
\begin{figure}[htbp]
		\centering
		\includegraphics[width=0.8\textwidth]{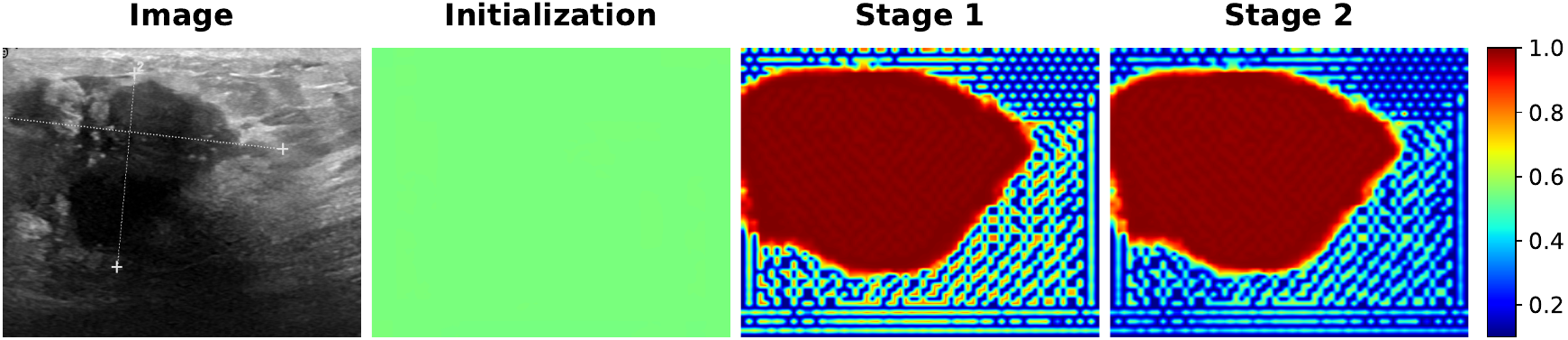} \\
		
		\includegraphics[width=0.8\textwidth]{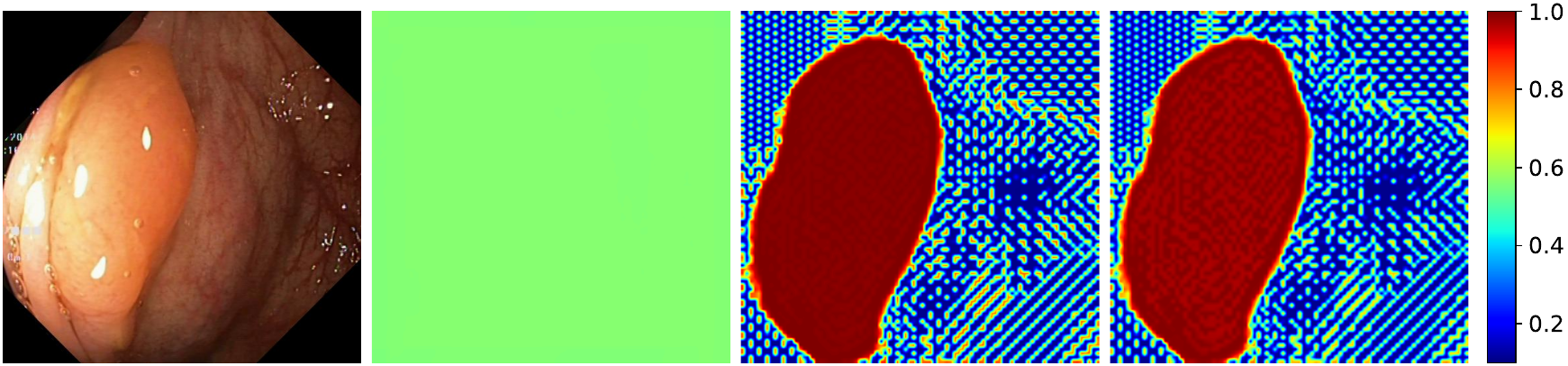} \\
		
		\includegraphics[width=0.8\textwidth]{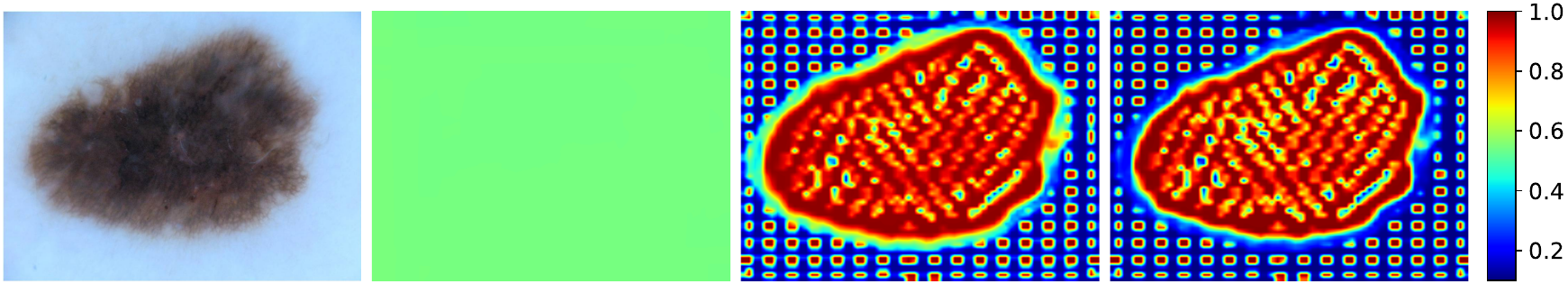} \\
		
		\caption{Evolution of the learned parameters $\mu$ across initialization and unfolding stages on the BUSI, Kvasir, and ISIC 2018 datasets (from top to bottom).}
		\label{fig:learned_mu}
	\end{figure}

	In our hyperbolic framework, the curvature coefficient $\mu$
	is no longer a fixed global constant but a spatially varying field $\mu(x,y)$ generated by a convolutional operator acting on the feature space $\mathbb{Z}$. The heatmaps in \cref{fig:learned_mu} illustrate its evolution from initialization to refined stages. Initially, the $\mu$ field is nearly uniform, though small variations exist due to its convolutional initialization from the feature space $\mathbb{Z}$. As the level‑set function $\phi$ evolves, $\mu$ becomes highly anisotropic, indicating that the network learns to assign local operator weights according to semantic structures. A consistent pattern across datasets is that high $\mu$ values (warm tones) are concentrated in homogeneous interior regions, where stronger curvature regularization suppresses noise and preserves topological continuity. Conversely, low $\mu$ values (cool tones) are localized along anatomical boundaries. By reducing curvature smoothing at these locations, the network allows data fidelity and deep prior forces to dominate, enabling precise boundary alignment. This adaptive modulation of $\mu$ thus can be viewed  as a learned edge‑stopping mechanism, ensuring both global smoothness and accurate delineation of fine anatomical details.
	
	A notable observation is the high consistency of parameter values between Stage 1 and Stage 2. This early stabilization indicates that LHMCF rapidly identifies an optimal physical regime for the hyperbolic flow. The convergence pattern explains why performance saturates at $K=2$, since once the physical parameters settle into a stable configuration, additional unfolding stages yield diminishing returns.
	
	\subsection{Evolution of the level-set function}
\begin{figure}[htbp]
		\centering
		\includegraphics[width=0.8\textwidth]{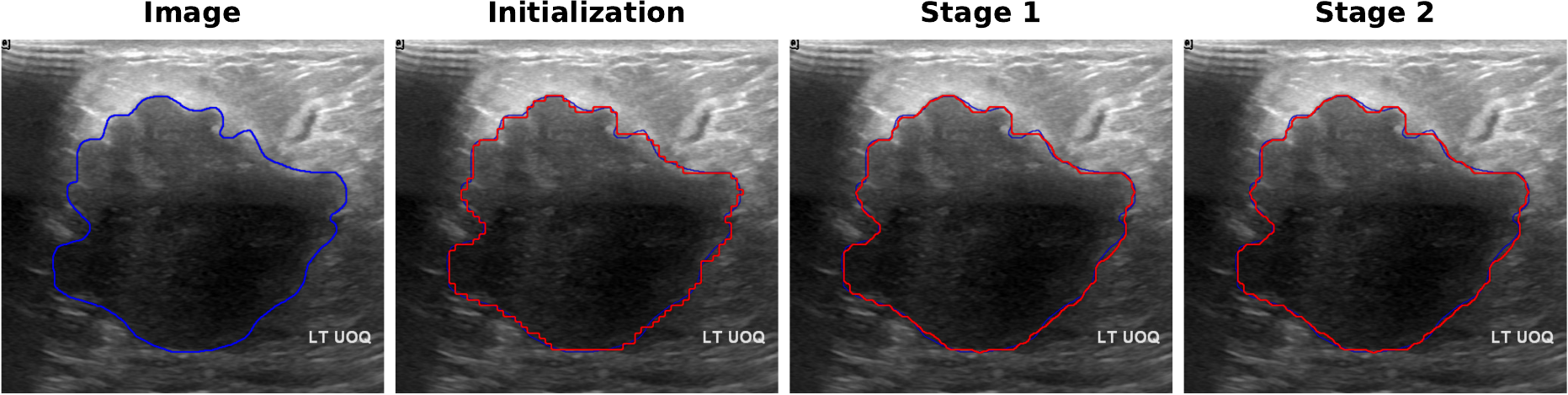}\\
		\vspace{0.1em}
		
		\includegraphics[width=0.8\textwidth]{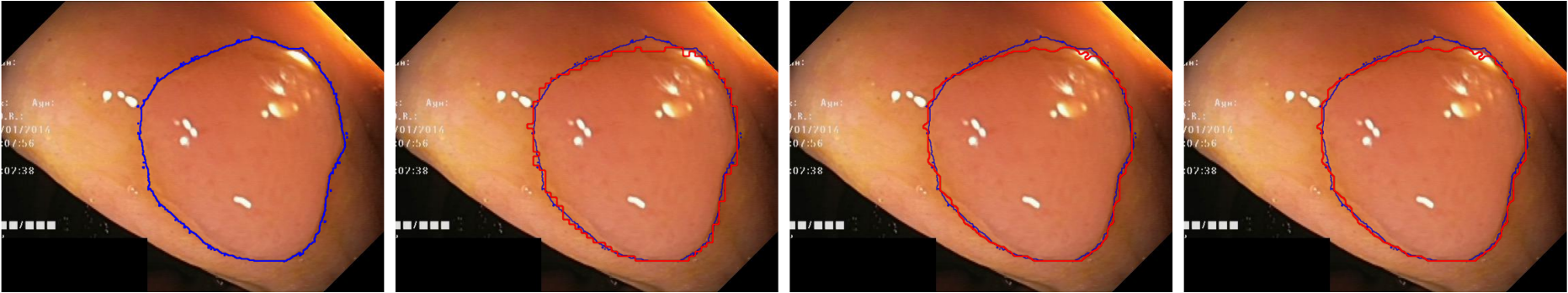} \\
		\vspace{0.1em}
		
		\includegraphics[width=0.8\textwidth]{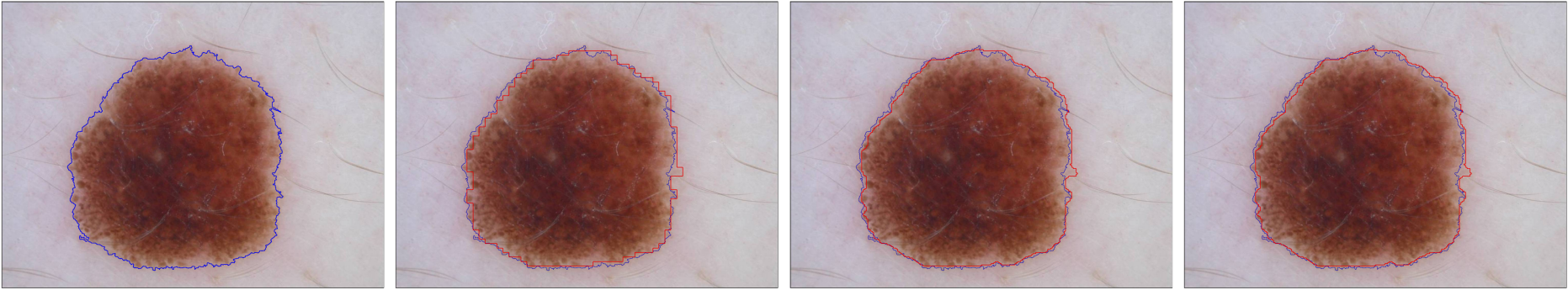} \\
		
		\caption{Evolution of the level-set function $\phi$ across initialization and unfolding stages on the BUSI, Kvasir, and ISIC 2018 datasets (from top to bottom). The blue contour denotes the ground-truth boundary, while the red contour represents the evolving level-set interface.}
		\label{fig:evolution_phi}
	\end{figure}

	To explore the dynamics of LHMCF‑Net, we visualize the evolution of the level-set function $\phi$ across unfolding stages. As illustrated in \cref{fig:evolution_phi}, the interface evolves from a semantically informed initialization toward the precise anatomical boundary. Unlike traditional variational models that rely on manual or random initializations, the initial contour $\phi_0$ generated by a convolutional mapping from the feature space $\mathbb{Z}$ already provides a well-localized estimate of the target structure. This initialization encodes high‑level semantic cues learned by the encoder, allowing the subsequent PDE evolution to focus on geometric refinement rather than coarse searching. In Stage 1, the red contour undergoes a rapid and coherent displacement toward the true boundary. This motion is governed by the synthesized force $F$, which integrates curvature-driven smoothing, feature space data fidelity, and the learned structural prior. During  Stage 2, the interface exhibits strong alignment with the ground truth. The contour becomes smoother in homogeneous regions while remaining sharply defined at boundary locations. This behavior reflects the effect of the learned anisotropic curvature coefficient $\mu(x,y)$, which suppresses regularization near edges and enforces topological coherence elsewhere. The high consistency between the final contour and the ground truth indicates that the learned physical parameters effectively steer the evolution toward a stable and accurate steady state within only two unfolding steps. Overall, these observations confirm that the proposed hyperbolic formulation yields a fast, stable, and interpretable evolution of the level‑set function.
	
	\subsection{Impact of initial velocity}
	Our numerical example here shows that if we choose the initial velocity properly, the second-order flow can behave significantly differently from the one with homogeneous initial velocity. This phenomenon fundamentally distinguishes hyperbolic dynamics from first‑order parabolic flows, where the evolution is entirely determined by the instantaneous gradient descent direction and is insensitive to initial momentum. To isolate this effect, by omitting the deep prior term  $\alpha R(\phi)$ in \Cref{eq:311} we consider the followng  purely damped quasilinear hyperbolic PDE
	\begin{displaymath}
		\phi_{tt}+\beta\phi_t=|\nabla\phi|\left(\mu\nabla\cdot\left(\frac{\nabla\phi}{|\nabla\phi|}\right)
		-\left(\lambda_1\|\mathbb{Z}-c_1\|_2^2-\lambda_2\|\mathbb{Z}-c_2\|_2^2\right)\right).
	\end{displaymath}
	We evaluate four representative initial velocity configurations: homogeneous velocity $v_0=0$, constant velocity $v_0=1$, oscillatory velocity $v_0=sinx\cdot siny$, and curvature‑adaptive velocity $v_0=-\gamma\kappa_0$, where $\gamma$ is a learned parameter, $\kappa_0$ denotes the curvature of the initial level‑set function $\phi_0$. As summarized in \cref{tab:velocity}, all inhomogeneous initial velocities improve segmentation accuracy compared with the standard zero velocity initialization. The curvature‑adaptive configuration achieves the most significant gains, reducing HD95 to 15.5806. These results suggest that injecting a momentum field aligned with the initial geometry helps the interface settle more precisely into the anatomical manifold, effectively mitigating the influence of noise‑induced local extrema.
	
	\begin{table}[htbp]
		\footnotesize
		\caption{Comparison of different initial velocity on the BUSI datasets. Best results are highlighted in bold.}\label{tab:velocity}
		\begin{center}
			\begin{tabular}{l l l l l l l} 
				\hline 
				Velocity & Acc $\uparrow$ & IoU $\uparrow$ & DSC $\uparrow$ & HD95 $\downarrow$ \\ 
				\hline 
				$v=0$ & 0.9699 & 0.7485 & 0.8513 & 17.7611 \\
				$v=1$ & 0.9727 & 0.7554 & 0.8552 & 17.4153 \\ 
				$v=sinxsiny$ & 0.9714 & 0.7574 & 0.8572 & 16.0380 \\ 
				$v=-\gamma\kappa_0$ & \textbf{0.9738} & \textbf{0.7601} & \textbf{0.8585} & \textbf{15.5806}\\ 
				\hline				
			\end{tabular}
		\end{center}
	\end{table}
	
	\section{Conclusion}
	\label{sec:conclusion}
	In this paper, we have proposed a hyperbolic mean curvature flow–based segmentation model (LHMCF) that integrates feature space data fidelity and deep structural priors into a second‑order dissipative hyperbolic PDE framework. Building upon this formulation, we further developed a lightweight deep unfolding network, LHMCF‑Net, which discretizes the coupled hyperbolic system into a sequence of learnable evolution stages. This unfolding strategy establishes a principled mapping between continuous physical dynamics and discrete neural operators, ensuring numerical stability, preserving physical interpretability, and enabling end‑to‑end optimization of evolutionary parameters. Experiments on three benchmark datasets demonstrate consistent improvements in accuracy, IoU, DSC, and HD95, highlighting the efficacy of the proposed approach in challenging medical images with low contrast and ambiguous boundaries. Additionally, the second‑order hyperbolic formulation also introduces a velocity field into the initial conditions, distinguishing it fundamentally from first‑order parabolic flows. While homogeneous initial velocity is often considered as the default, our experiments intuitively reveal that inhomogeneous initial velocities can lead to markedly different segmentation behaviors, suggesting that the hyperbolic flow possesses a richer dynamical structure than previously recognized. These observations indicate that the proposed LHMCF framework not only contributes a novel PDE‑driven segmentation method but also raises interesting research questions for the fields of PDE and geometric analysis. In particular, for inhomogeneous initial velocities, fundamental well‑posedness problems remain analytically open. Investigating the convergence properties and stability of hyperbolic geometric flows under general initial velocity conditions will be an important direction for future research.
	
	\section*{Acknowledgments}
	The authors S. Duan and S. Huang would like to thank Professor Chunlin Wu and Professor Shihui Ying for helpful discussions on unfolding algorithms. In addition, S. Duan thanks Dr. Shengdong Zhang for providing the codes of LMS-Net in \cite{zhang2025lms}.
	
	\section*{Code Availability}
	Codes will be available at  https://github.com/duanss0517/LHMCF-Net.

	\bibliographystyle{plain}
	\bibliography{references}
	
\end{document}